\PassOptionsToPackage{dvipsnames}{xcolor}
\PassOptionsToPackage{table}{xcolor}
\PassOptionsToPackage{xcdraw}{xcolor}
\PassOptionsToPackage{most}{tcolorbox}
\documentclass[11pt]{article}

\usepackage[final]{acl}
\usepackage{times}
\usepackage{latexsym}

\usepackage[T1]{fontenc}

\usepackage[utf8]{inputenc}

\usepackage{microtype}

\usepackage{inconsolata}

\usepackage{graphicx}

\usepackage{hyperref}       
\usepackage{url}            
\usepackage{booktabs}       
\usepackage{amsfonts}       
\usepackage{nicefrac}       
\usepackage{microtype}      
\usepackage{xcolor}         
\usepackage{lipsum}
\usepackage{xspace}
\usepackage{romannum}
\usepackage{graphicx,pifont}
\usepackage{multirow}
\usepackage{caption} 
\usepackage{amssymb}
\usepackage{amsmath}
\usepackage{tabularray}
\usepackage{lipsum}
\usepackage{subcaption}
\usepackage{tcolorbox}
\usepackage{algorithm}
\usepackage{algpseudocode}
\usepackage{balance}

\colorlet{shadecolor}{gray!40}

\definecolor{darkgreen}{RGB}{1,50,32}
\definecolor{deepskyblue}{rgb}{0.0, 0.75, 1.0}
\definecolor{cvprblue}{rgb}{0.21,0.49,0.74}
\definecolor{electriclime}{rgb}{0.8, 1.0, 0.0}
\definecolor{ferrarired}{rgb}{1.0, 0.11, 0.0}

\newcommand{\ourapproach}{\textsc{\textbf{RAVEN}}\xspace}
\newcommand{\dname}{\textbf{AVS-QA}\xspace}
\newcommand{\amodule}{\textbf{QuART}\xspace}

\newcommand{\xmark}{\textcolor{OrangeRed}{\ding{55}}}
\newcommand{\tmark}{\textcolor{ForestGreen}{\ding{51}}}

\newcommand{\parlabel}[1]{\noindent\textbf{#1}.}

\newcommand{\fres}[3]{\xspace${#1}_{\scalebox{.8}{\textcolor{#3}{{#2}\%}}}$}

\newcommand{\frestwo}[3]{\xspace${#1}_{\scalebox{.8}{\textcolor{#3}{{#2}}}}$}

\newcommand\blfootnote[1]{%
  \begingroup
  \renewcommand\thefootnote{}\small{\footnote{#1}}%
  \addtocounter{footnote}{-1}%
  \endgroup
}

\title{~\includegraphics[height=30pt]{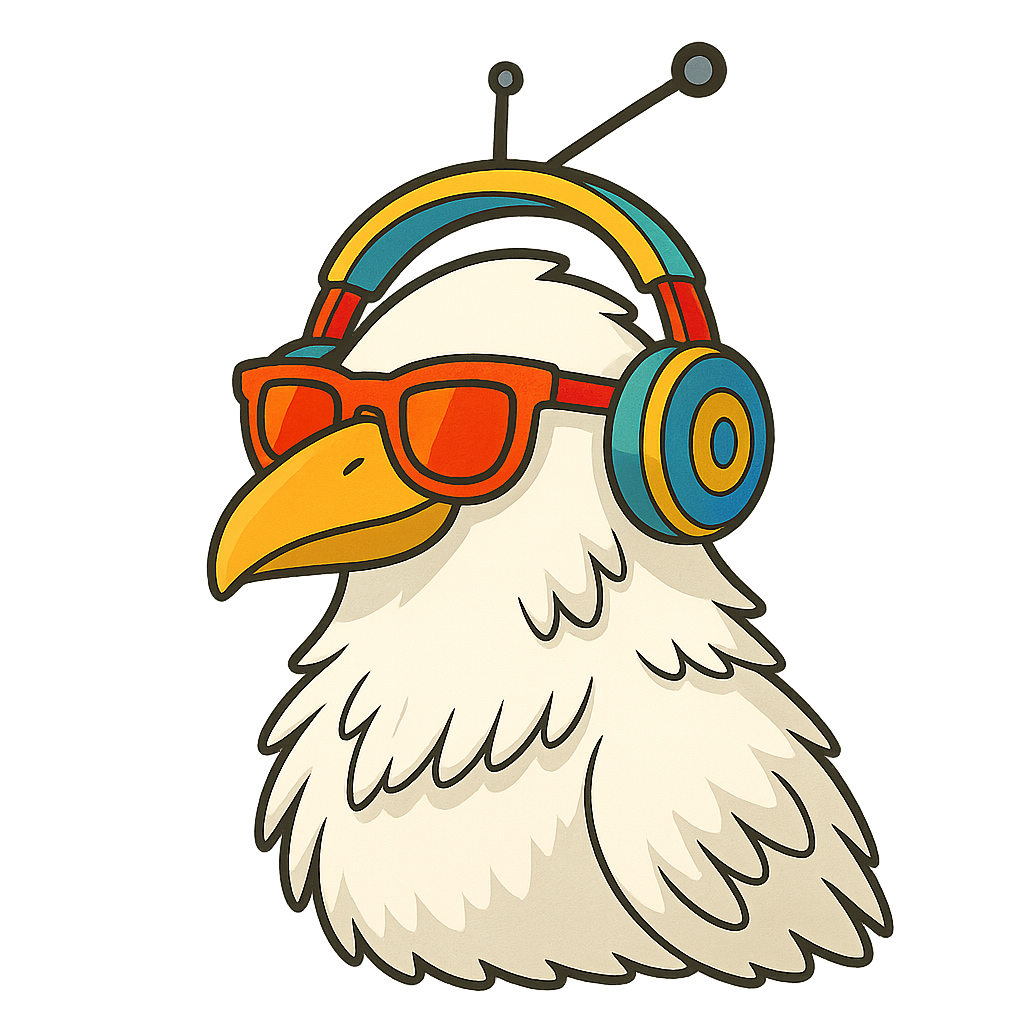}\ourapproach: 
Query-Guided \underline{R}epresentation Alignment for Question Answering over \underline{A}udio, \underline{V}ideo, \underline{E}mbedded Sensors, and \underline{N}atural Language
}
\author{%
  Subrata Biswas\textsuperscript{*},  Mohammad Nur Hossain Khan\textsuperscript{*}, Bashima Islam\\
  Department of Electrical \& Computer Engineering\\
  Worcester Polytechnic Institute\\
  Worcester, MA 01609 \\
  \texttt{\{sbiswas, mkhan, bislam\}@wpi.edu}
}

\begin{document}
\maketitle

\begin{abstract}
Multimodal question answering (QA) often requires identifying which video, audio, or sensor tokens are relevant to the question. Yet modality disagreements are common: off-camera speech, background noise, or motion outside the field of view often mislead fusion models that weight all streams equally.
We present \ourapproach, a unified QA architecture whose core is \amodule, a query-conditioned cross-modal gating module that assigns scalar relevance scores to each token across modalities, enabling the model to amplify informative signals and suppress distractors before fusion. \ourapproach is trained through a three-stage pipeline comprising unimodal pretraining, query-aligned fusion, and disagreement-oriented fine-tuning -- each stage targeting a distinct challenge in multi-modal reasoning: representation quality, cross-modal relevance, and robustness to modality mismatch.
To support training and evaluation, we release \dname, a dataset of 300K synchronized Audio--Video-Sensor streams paired with automatically generated question-answer pairs.
Experimental results on seven multi-modal QA benchmarks -- including egocentric and exocentric tasks -- show that \ourapproach achieves up to 14.5\% and 8.0\% gains in accuracy compared to state-of-the-art multi-modal large language models, respectively. Incorporating sensor data provides an additional 16.4\% boost, and the model remains robust under modality corruption, outperforming SOTA baselines by 50.23\%. Our code and dataset are available at \url{https://github.com/BASHLab/RAVEN}.

\end{abstract}
\blfootnote{\textsuperscript{*} These authors contributed equally.}
\vspace{-2em}
\section{Introduction}
Answering natural language questions in multimodal settings often requires reasoning over visual, auditory, and sensor inputs to extract the most relevant evidence \cite{wanniarachchi2025mimic}. Yet real-world signals are rarely clean or aligned: off-camera speech, background noise, and unobserved motion can introduce conflicts across modalities. Without identifying which inputs are relevant to the question, fusion models may attend to irrelevant signals and overlook critical evidence.

We introduce \ourapproach, a unified architecture for question answering over video, audio, and sensor inputs. It resolves cross-modal conflicts by reasoning about modality relevance. At its core is \amodule, a query-conditioned cross-modal gating module that assigns scalar relevance scores to each token. These scores suppress distractors and amplify informative signals before fusion, enabling the model to produce context-sensitive representations grounded in the question.

\begin{figure*}
    \centering    \includegraphics[width=\textwidth]{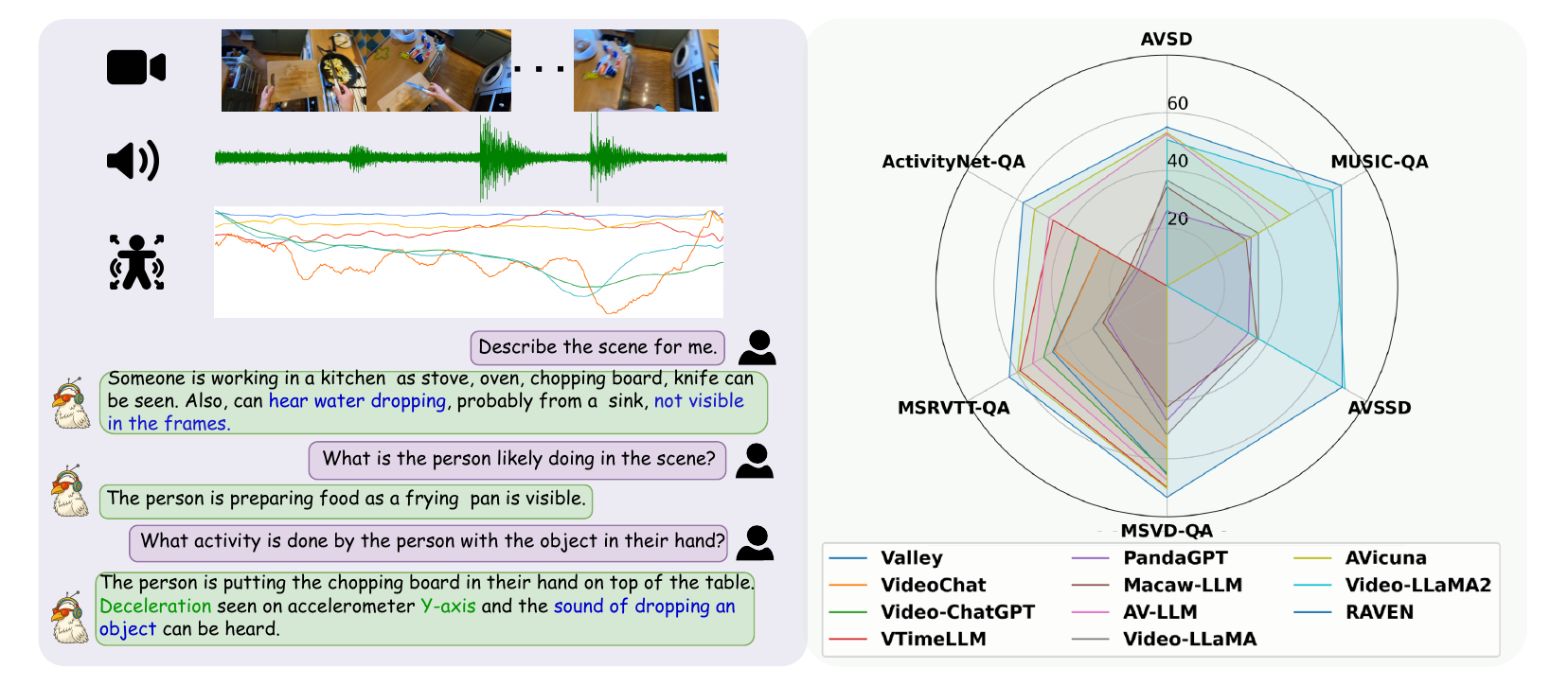}
    \vspace{-2em}
    \caption{\ourapproach jointly interprets video, audio, and sensor signals (e.g., inertial measurement unit or IMU) to answer fine-grained, context-aware questions. 
    It outperforms existing MLLMs across six QA benchmarks, demonstrating robust generalization through multi-modal alignment.}
    \vspace{-1em}
    \label{fig:teaser}
\end{figure*}

This challenge intensifies with sensor data integration. Unlike visual and auditory streams, sensor inputs capture latent physical dynamics, such as acceleration, orientation, and velocity, but often arrive asynchronously, are noisy, and lack semantic anchors. Their relevance also varies by question. For instance, when asked \textit{``Did the user place the object gently?''}, only audio (e.g., impact sound) and motion traces (e.g., deceleration) are informative, while visual frames may mislead. \amodule's query-conditioned filtering allows the model to focus on such signals while ignoring irrelevant tokens. Figure~\ref{fig:teaser} illustrates this behavior and highlights the resulting performance gains.

Recent advances in multimodal large language models (MLLMs) have enabled perception-language reasoning by combining pretrained LLMs with modality-specific encoders and fusion strategies~\cite{llava, videollava, qwenaudio}. Models such as Flamingo~\cite{openflamingo}, Video-LLaMA~\cite{videollama1}, and AVicuna~\cite{avicuna} have achieved strong results on video captioning, video QA, and audio-language tasks~\cite{li2023seed, yu2023mm, liu2024mmbench}. However, these systems typically focus on vision and audio, ignoring embedded sensor modalities that are critical in domains like AR/VR, robotics, and mobile health. Moreover, they often assume clean, synchronized inputs and rely on projection, cross-attention~\cite{ye2024x, wu2024next, chowdhury2025egoadapt}, or contrastive alignment~\cite{clip, clap} —approaches that break down under modality misalignment. In contrast, \ourapproach uses query-conditioned token-level filtering via \amodule\ to dynamically attend to the most informative modality stream at each timestep.

We train \ourapproach using a three-stage pipeline: (1) unimodal pretraining to improve encoder specialization, (2) query-aligned fusion to teach relevance modeling, and (3) disagreement-oriented fine-tuning to increase robustness under modality mismatch. Each stage addresses a distinct challenge in multimodal reasoning, yielding an average $26.87\%$ improvement over training without disagreement-oriented fine-tuning.

To support training and evaluation, we release \dname, a dataset of 300K automatically generated \texttt{\{Audio, Video, Sensor, QA\}} quadruples from egocentric scenarios. To our knowledge, it is the first large-scale QA benchmark with synchronized input streams and question–answer supervision across all three modalities (See Table~\ref{tab:benchmark_comparison}). 

\ourapproach, powered by \amodule, achieves state-of-the-art performance on seven QA benchmarks, with gains of up to 14.5\% over VideoLLaVA~\cite{videollava} and 8.0\% over AVicuna \cite{avicuna}  on egocentric and exocentric tasks, respectively. Incorporating sensor data yields an additional 16.4\% boost, and under modality corruption, \ourapproach retains a 50.23\% improvement over prior systems-demonstrating robust, query-aware reasoning across diverse multimodal inputs.
We summarize our contributions below:

\begin{table}[!htb]
\centering
\caption{Comparison of egocentric QA benchmarks. \dname is the only dataset with all three modalities, four QA types, and large-scale automated supervision.
}
\label{tab:benchmark_comparison}
\resizebox{\linewidth}{!}{
\begin{tabular}{llllllll}
\toprule[2pt]
\textbf{Benchmark}     & \textbf{A} &  \textbf{V}   &  \textbf{S}  & \textbf{\begin{tabular}[c]{@{}c@{}}Data \\ Source\end{tabular}}         & \textbf{\begin{tabular}[c]{@{}c@{}}Answer \\ Type\end{tabular}} & \textbf{Evaluator}       & \textbf{Size} \\ \midrule[1.5pt]

EgoTaskQA       & \tmark & \tmark & \xmark  & \begin{tabular}[c]{@{}c@{}}Crowd- \\ sourcing\end{tabular}    & OE               & \begin{tabular}[c]{@{}c@{}}Crowd- \\ sourcing\end{tabular}    & 40K           \\
\arrayrulecolor{shadecolor} \cmidrule[1pt](l){1-8}\arrayrulecolor{black}
EgoVQA             & \tmark &  \tmark & \xmark            & Handcraft              & MC                   & Accuracy             & 520           \\
\arrayrulecolor{shadecolor} \cmidrule[1pt](l){1-8}\arrayrulecolor{black}
EgoThink                & \tmark & \tmark & \xmark        & Handcraft              & OE                   & LLMs                 & 700           \\
\arrayrulecolor{shadecolor} \cmidrule[1pt](l){1-8}\arrayrulecolor{black}
VidEgoThink   &  \tmark & \tmark & \xmark                   & \begin{tabular}[c]{@{}c@{}}Egocentric \\ video\end{tabular}            & OE                   & LLMs                     & 1.2K         \\
\arrayrulecolor{shadecolor} \cmidrule[1pt](l){1-8}\arrayrulecolor{black}
MM-Ego              & \tmark & \tmark & \xmark                     & \begin{tabular}[c]{@{}c@{}}Multimodal\\ (AV)\end{tabular}             & OE / MC              & \begin{tabular}[c]{@{}c@{}}Accuracy, \\ LLMs /CE\end{tabular}          & 10K           \\
\arrayrulecolor{shadecolor} \cmidrule[1pt](l){1-8}\arrayrulecolor{black}
\dname      & \tmark & \tmark & \tmark                     & \begin{tabular}[c]{@{}c@{}}Egocentric \\ video\end{tabular}  &  \begin{tabular}[c]{@{}c@{}}MC / OE \\ TF /CE\end{tabular}             & LLMs          & 300K        \\
\bottomrule[2pt]
\end{tabular}}
\vspace{-1em}
\end{table}

\noindent$\bullet$ We propose \ourapproach, a unified QA model that integrates video, audio, and sensor inputs using \amodule, a query-conditioned gating module to filter distractors before fusion

\noindent$\bullet$ Introduction of query-aligned fusion and disagreement-oriented fine-tuning after unimodal pre-training enhances representation, relevance, and robustness to cross-modal disagreement.

\noindent$\bullet$ We release \dname, a 300K-sample dataset with synchronized audio, video, sensor streams, and auto-generated QA pairs.

\noindent$\bullet$ We achieve state-of-the-art results on seven benchmarks, with strong performance across egocentric, exocentric, and corrupted-input settings.
\begin{figure*}
    \centering
    \includegraphics[width=\linewidth]{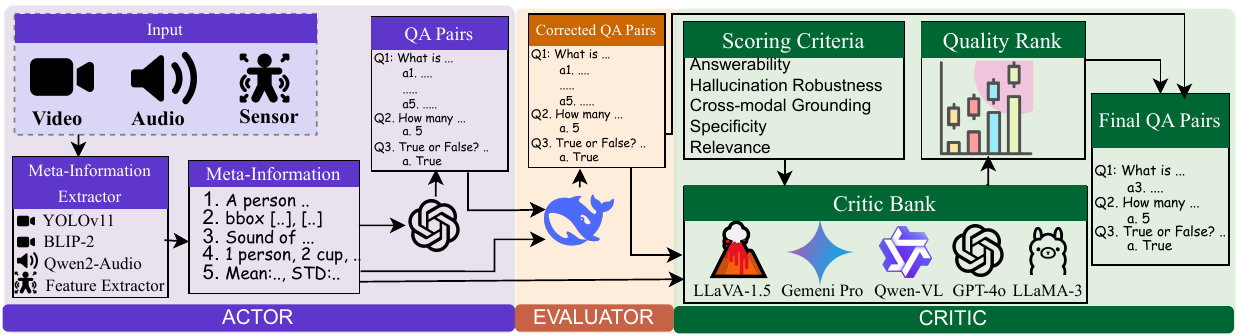}
    \vspace{-1.8em}
    \caption{Overview of the \dname dataset pipeline. Given synchronized audio–video–sensor input, the Actor generates metadata and QA pairs, the Evaluator filters weakly grounded examples, and the Critic ranks quality across five axes. The process is fully automated and yields 300K high-quality QA examples across four types.}
\vspace{-1em}\label{fig_dataset_curation}
\end{figure*}
\section{Related Work}
\parlabel{Large and Multi-modal Language Models}
Large language models (LLMs) such as LLaMA~\cite{llama} and GPT-4~\cite{gpt4} have demonstrated strong reasoning abilities. Multi-modal language models (MLLMs) extend LLMs with modality-specific encoders and fusion modules for visual or auditory inputs~\cite{blip2, llava, qwenvl, valley, qwen2audio, audioflamingo}. Representative models such as Flamingo~\cite{alayrac2022flamingo}, LLaVA~\cite{llava}, and Video-LLaMA~\cite{videollama1} achieve impressive results on vision-language and audio-video QA through instruction tuning. However, these systems typically ignore embedded sensor modalities and assume synchronized, clean inputs. Sensor-aware models--such as LLMSense~\cite{llmsense}, IMUGPT~\cite{imugpt}, and OpenSQA/LLASA~\cite{llasa}--process inertial signals in isolation, without visual or auditory grounding. ImageBind~\cite{imagebind} supports multiple modalities but lacks QA supervision or cross-modal reasoning. In contrast, our framework performs query-guided alignment across video, audio, and sensor inputs with direct QA grounding. See Appendix~\ref{supp_related_works} for full citations.

\parlabel{Multi-modal Feature Alignment}
Token-level fusion across modalities is central to MLLM performance. Dual encoders like CLIP~\cite{clip} and fusion-based models such as LLaVA~\cite{llava} and Q-Former~\cite{blip2} align vision and language. Extensions like Hierarchical Q-Former~\cite{azad2025hierarq}, Smaug~\cite{lin2023smaug}, and MACAW~\cite{macawllm} adapt this to temporal signals but are optimized for audio-visual tasks. These approaches struggle under sensor-specific noise, asynchrony, or modality mismatch. Our proposed \amodule assigns query-conditioned scalar weights to cross-modal tokens, enabling selective fusion and robust reasoning under disagreement.

\parlabel{Multi-modal Datasets}
Existing corpora support audio-visual (e.g., HowTo100M~\cite{panda70m}, AudioCaps~\cite{audiocaps}) and image-language learning (e.g., CC3M~\cite{cc3m}). QA-focused datasets such as AVQA~\cite{avqa}, MusicAVQA~\cite{musicavqa}, and MSRVTT-QA~\cite{msrvttqa} do not include sensor data. Egocentric QA datasets like Ego4D~\cite{ego4d} and EgoTaskQA~\cite{jia2022egotaskqa} lack synchronized video-audio-sensor input. To address this, we introduce \dname, a 300K-example dataset of {audio, video, sensor, QA} quadruples with synchronized streams, four question types, and frame-level alignment. Table~\ref{tab:benchmark_comparison} summarizes its scope.

\section{AVS-QA: Multi-Modal Dataset Curation Pipeline}

Despite rapid progress in multi-modal QA, no existing benchmark provides aligned supervision across video, audio, and sensor inputs. Prior QA datasets are either limited to vision-language pairs or omit sensor signals entirely (see Table~\ref{tab:benchmark_comparison}). To bridge this gap, we introduce \dname, a dataset of 300K automatically generated \{\texttt{video, audio, sensor, QA}\} quadruples. This scale exceeds the combined size of existing egocentric QA datasets by a factor of four. Unlike prior work, \dname includes four question types--open-ended (OE), closed-ended (CE), multiple-choice (MC), and true/false (TF)--supporting both generative and retrieval-style evaluation.

\dname is constructed via a fully automated, three-stage Actor--Evaluator--Critic pipeline, illustrated in Figure~\ref{fig_dataset_curation}. The pipeline takes as input a multi-modal triplet \( \mathcal{D} = (v, a, s) \), where \( v \), \( a \), and \( s \) denote temporally aligned video, audio, and sensor streams, and produces question-answer pairs \( (q, A) \in \mathcal{Q} \). Formally, the dataset generation process is defined as a mapping function \( F: \mathcal{D} \rightarrow \mathcal{Q} \), yielding synchronized \(\{v, a, s, q, A\}\) tuples.

\parlabel{Actor: Multi-modal Prompt Generation}
The Actor constructs an enriched scene description \( \mathcal{M} \) from each triplet \( \mathcal{D} \). 
We extract visual features using BLIP-2~\cite{blip2} (frame captioning) and YOLOv11~\cite{yolov11} (object detection, and localization); audio features using Qwen2-Audio-7B~\cite{qwen2audio} (transcription and event labels); and sensor features using a 200 Hz statistical extractor~\cite{llasa} over 15-second IMU windows (e.g., mean, RMS, skewness).
These cues are concatenated into a natural language prompt, from which the Actor generates four QA types: open-ended, closed-ended, multiple-choice, and true/false. For open-ended questions, five candidate answers are produced for filtering, and one final answer is retained.

\parlabel{Evaluator: Modality-Consistency Filtering}
Given a candidate QA pair \( (q, A) \) generated from meta-information \( \mathcal{M} \), the Evaluator verifies that the referenced modality or modalities are supported by the corresponding input triplet \( (v, a, s) \in \mathcal{D} \). For instance, motion-related questions require significant activity in the sensor stream (e.g., variance spike), while visual or auditory references must align with detected objects or acoustic summaries. Pairs lacking sufficient grounding are discarded. To ensure diversity, the Evaluator enforces a balanced mix of single- and cross-modality QA types.

\parlabel{Critic: Quality Ranking via LLM Scoring}
For each candidate pair, the Critic applies an ensemble of instruction-tuned LLMs to assess QA quality. Inspired by LLM-as-judge paradigms~\cite{fu2023gptscore, zheng2023judging}, we define a quality vector \( \mathcal{C}(q, A) = [s_1, s_2, s_3, s_4, s_5] \in \mathbb{R}^5 \), where each score corresponds to one of five axes: \textit{answerability}, \textit{hallucination robustness}, \textit{modality grounding}, \textit{specificity}, and \textit{semantic relevance}. A QA pair is discarded if any component score falls below a task-specific threshold (See Appendix~\ref{supp_dataset_curation}). This stage ensures that all retained examples are interpretable, grounded, and semantically meaningful. The final dataset contains short-form answers across four formats (open-ended, closed-ended, multiple-choice, and true/false), supporting both retrieval and generation in most formats.


\parlabel{Output}
\dname is built from egocentric clips in Ego4D~\cite{ego4d} and EPIC-Kitchens-100~\cite{epic-kitchen}, with each example containing synchronized video, audio, sensor data, and a verified answer. The dataset spans 300K QA pairs across three modalities, four QA types, and dual perspectives--offering diverse, fine-grained supervision for multi-modal reasoning. We randomly selected 300 samples from the dataset and conducted a human evaluation following the criteria described in Appendix~\ref{human_evaluation}. Additional statistics and details are provided in Appendix~\ref{supp_dataset_curation}. For privacy and ethical considerations, see Section~\ref{ethical_statement}. The \dname dataset has been publicly released under CC 4.0 license to support reproducibility.

\begin{figure*}
    \centering
    \includegraphics[width=1\textwidth]{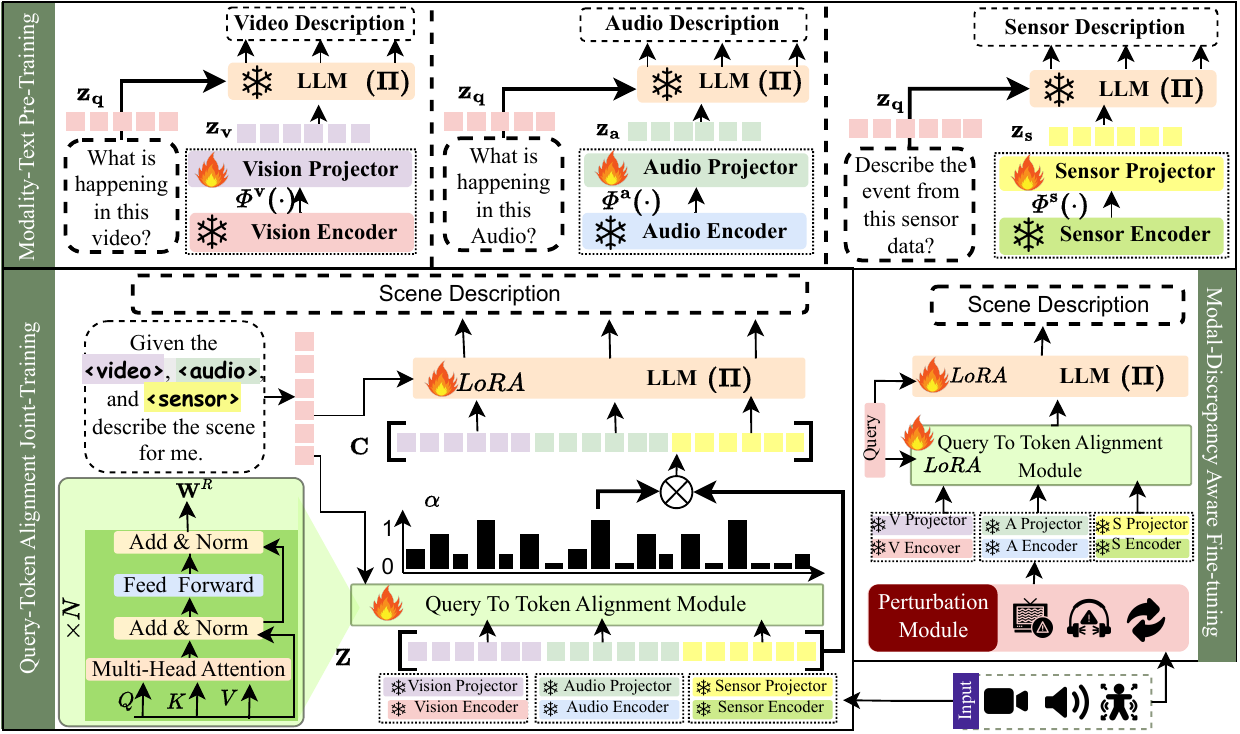}
    \vspace{-2em}
    \caption{Overview of \ourapproach. Each modality (video, audio, sensor) is encoded using pretrained encoders and projected into a shared space. The \amodule module performs query-conditioned token relevance scoring to align informative tokens across modalities. The figure also highlights the three-stage training pipeline for alignment-aware multi-modal reasoning. 
    Here, \includegraphics[height=9pt]{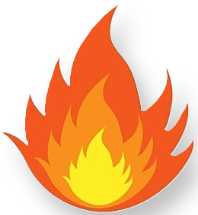} and \includegraphics[height=9pt]{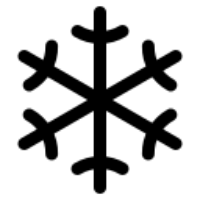} represent trainable and frozen components, respectively.
    }
    \vspace{-1em}
    \label{fig_model_arch}
\end{figure*}
\section{\ourapproach Framework: Query-Token Alignment for Multi-Modal Fusion}
\label{framework}

\ourapproach performs query-conditioned fusion of video, audio, and sensor inputs via token-level alignment. As shown in Figure~\ref{fig_model_arch}, inputs from each modalities are processed through individual pretrained encoders and projected to a shared space. Our core module, \amodule (Query-Aligned Representation of Tokens), computes query-aware relevance scores across all modalities, enabling robust reasoning under noisy or misaligned inputs. We describe each component below and architecture, training, and implementation details available in Appendix~\ref{supp_arch} and ~\ref{supp:train_implement}.


\parlabel{Modality-Specific Feature Encoders}
Given a triplet \( \mathcal{D} = \{v, a, s\} \), each modality is encoded and projected to \( \mathbb{R}^{L_m \times E} \). Video frames \( v = \{I_t\}_{t=1}^T \) are sampled uniformly and encoded using \texttt{SigLIP-so-400m}~\cite{siglip}, yielding \( \mathbf{z}_v = \Phi^v(v) \in \mathbb{R}^{L_v \times E} \). Audio is transformed into a Kaldi-fbank spectrogram~\cite{povey2011kaldi} and encoded via \texttt{BEATs}~\cite{beats} to obtain \( \mathbf{z}_a = \Phi^a(a) \in \mathbb{R}^{L_a \times E} \). Sensor data--multi-axis IMU streams--are encoded using \texttt{LIMU-BERT}~\cite{limubert}, producing \( \mathbf{z}_s = \Phi^s(s) \in \mathbb{R}^{L_s \times E} \) (See Appendix~\ref{supp_ablation_study} for ablation).

\parlabel{Language Decoder and Query Embedding} 
We use \texttt{Qwen2-7B-Instruct}~\cite{qwen2} as the decoder-only language model \( \Pi \). Its tokenizer maps the query \( Q \) to token embeddings \( \mathbf{z}_q \in \mathbb{R}^{L_q \times E} \). 
Each modality encoder--\( \Phi^v(v) \), \( \Phi^a(a) \), \( \Phi^s(s) \)--is followed by a projection layer that projects extracted feature into the shared space \( \mathbb{R}^{L_m \times E} \). For simplicity, \( \Phi^m(\cdot) \) refers to the combined encoder and projection for modality \( m \in \{v, a, s\} \) (See Appendix~\ref{supp_projection_layer}).

\parlabel{\amodule: Query-Aligned Representation of Tokens} The \amodule module performs query-conditioned token selection over multi-modal inputs.
Given visual, audio, and sensor token sequences \( \mathbf{z}_v, \mathbf{z}_a, \mathbf{z}_s \in \mathbb{R}^{L_m \times E} \), we concatenate them into a unified token matrix \( \mathbf{Z} \in \mathbb{R}^{L \times E} \), where \( L = L_v + L_a + L_s \). 
We apply multi-head attention between the query embedding \( \mathbf{z}_q \) and \( \mathbf{Z} \) as:
$
\mathbf{Q} = \mathbf{z}_q \mathbf{W}^Q, \quad \mathbf{K} = \mathbf{Z} \mathbf{W}^K, \quad \mathbf{V} = \mathbf{Z} \mathbf{W}^V$,
where \( \mathbf{W}^Q, \mathbf{W}^K, \mathbf{W}^V \in \mathbb{R}^{E \times d_k} \) are learned projections. Temporal order is preserved via sinusoidal positional embeddings, as in standard Transformer encoders.
The aggregated attention output is 
$\mathbf{M} = \texttt{softmax}\left( \frac{\mathbf{Q} \mathbf{K}^\top}{\sqrt{d_k}} \right) \mathbf{V}$. 

Unlike standard multi-head attention--which uses similarity-based weights across modalities--\amodule introduces a relevance projection head, \( \mathbf{W}^R \in \mathbb{R}^{E \times L} \), that learns to score tokens conditioned on the query. 
This separation enables the model to prioritize semantically relevant tokens even when distractors receive high attention weights--a key advantage under modality mismatch. 
\amodule uses learned relevance scores to prioritize tokens based on the question. For instance, when asked about gentle placement, it emphasizes sensor deceleration and impact sounds while down-weighting static visual frames. If the camera is occluded and the user trips, only IMU spikes and audio thuds are informative--QuART gates out blank video. This behavior generalizes, suppressing off-screen audio when questions target visual actions.
This token-level relevance scores are computed as:
$\boldsymbol{\alpha} = \texttt{softmax}(\mathbf{M} \mathbf{W}^R)$.
The fused context vector, $\mathbf{C} = \sum\nolimits_{j=1}^{L} \alpha_j \mathbf{Z}_j$ aggregates query-weighted tokens across all modalities and conditions the LLM decoder. 
This learned relevance outperforms raw attention (Section~\ref{ablation_study}).



\parlabel{Training Objective} The decoder \( \Pi \) predicts the output sequence \( \{y_t\}_{t=1}^{T} \) conditioned on \( \mathbf{C} \), trained via autoregressive cross-entropy:
$\mathcal{L}_{\amodule} = - \frac{1}{T} \sum\nolimits_{t=1}^{T} \log p_\theta(y_t \mid y_{<t}, \mathbf{C})
$.
To promote sparse selection of relevant tokens, we introduce an entropy-based regularizer:
$\mathcal{L}_{\text{reg}} = \sum\nolimits_{j=1}^{L} \alpha_j \log \alpha_j$.The total loss is 
\vspace{-0.75em}
\begin{equation}
\label{eq:loss}
  \mathcal{L}_{\ourapproach} = \mathcal{L}_{\amodule} + \lambda \mathcal{L}_{\text{reg}}  
  \vspace{-0.5em}
\end{equation}
We encourage sparsity via entropy regularization scaled by $\lambda$. Relevance is disabled in early stages $(\mathbf{C} = \mathbf{Z}, \lambda = 0)$ and enabled in the final stage with $\lambda = 0.001$. 
See Appendix~\ref{supp:train_implement} for implementation \& hyperparameters and Appendix~\ref{supp_cost} for cost analysis. Table~\ref{tab:wr} and Appendix~\ref{supp_ablation_study} demonstrate \amodule's advantage over SOTA alignment methods.

\section{Alignment-Aware Multi-Stage Training for Multi-Modal Reasoning}
\label{alignment}
We adopt a three-stage training procedure to optimize \ourapproach and its query-conditioned alignment module. Each stage targets a distinct component--projection alignment, query-token fusion, and robustness to input degradation--stabilizing learning and reducing cross-modal interference (Figure~\ref{fig_model_arch}).

\parlabel{Stage \Romannum{1}: Modality-Text Pre-Training}
In this pretraining stage, we use a large-scale, weakly labeled dataset of modality-text pairs: \texttt{\{video, text\}}, \texttt{\{image, text\}}, \texttt{\{audio, text\}}, and \texttt{\{sensor, text\}}, collected from caption-rich sources, e.g., WavCaps~\cite{wavcaps}, and InternVid-10M~\cite{internvid10m}. We adopt a sequential, modality-specific training strategy to avoid inter-modal interference and stabilize projection learning. Supervision is provided via natural language captions or transcriptions paired with raw modality inputs, such as video subtitles, audio narrations, and wearable sensor logs.
For each modality \( m \in \{v, a, s\} \), we freeze the pretrained encoder \( \Phi^m(\cdot) \) and language model \( \Pi \), and update only the corresponding projection head \( P^m \) to align with textual supervision. All three branches are trained in succession using the same LLM decoder, promoting consistent language grounding across modalities.

\parlabel{Stage \Romannum{2}: Query-Token Alignment Joint-Training}
After modality-specific alignment, we train the QuART module to perform token-level fusion conditioned on natural language queries. We use the AVS-QA dataset for this stage, which provides synchronized video, audio, sensor, and query-answer supervision (Equation~\ref{eq:loss}).
All modality encoders \( \Phi^v, \Phi^a, \Phi^s \) and their projection heads are frozen to preserve previously learned alignments. We initialize \amodule from scratch and train it to compute relevance-weighted token representations that bridge cross-modal information and the query context. In parallel, we fine-tune the LLM decoder \( \Pi \) using Low-Rank Adaptation (LoRA)~\cite{lora} with rank 256, offering efficient adaptation to fused multi-modal inputs without catastrophic forgetting.
This stage enables query-aware modality fusion, teaching \ourapproach to prioritize informative tokens for reasoning and generation.

\parlabel{Stage \Romannum{3}: Modal-Discrepancy Aware Fine-tuning}
To improve robustness under real-world conditions, we fine-tune \ourapproach using perturbed multi-modal inputs that simulate modality mismatch--such as dropped sensor packets or off-screen audio. We apply stochastic transformations independently to each modality: video undergoes frame jitter, dropout, or temporal inversion; audio is corrupted with Gaussian noise, reversed, or replaced with unrelated samples; sensor signals are perturbed with zero-centered Gaussian noise based on empirical variance (see Appendix~\ref{supp_modal_dicrepency}).
Perturbed inputs \( \tilde{\mathcal{D}} = \{\tilde{v}, \tilde{a}, \tilde{s}\} \) are encoded by frozen encoders \( \Phi^m \) and passed through the trained QuART module and LoRA-adapted decoder \( \boldsymbol{\Pi} \). During this stage, we activate entropy regularization to sharpen token relevance and encourage sparse, discriminative alignment. We set $\lambda = 0.001$ in the final stage, as it yields the best trade-off between sparsity and accuracy (see Section~\ref{ablation_study}); earlier stages use $\lambda = 0.$
See Appendix~\ref{supp:train_implement} for full training details.

\begin{table*}[!htb]
\caption{Comparison of \ourapproach and prior MLLMs on \textbf{exocentric} open-ended video QA (MSVD-QA, MSRVTT-QA, ActivityNet-QA) and audio-visual QA (AVSD, MUSIC-QA) benchmarks. Best and second-best scores are in \textbf{bold} and \underline{underline}. $^*$ indicates scores reproduced by us.}
\vspace{-0.5em}
\label{tab_av_main_result_exo}
\centering
\resizebox{\linewidth}{!}{
\begin{tabular}{l|cc|rr|cccccc}
\toprule[2pt]
\multicolumn{1}{c|}{} &
  \multicolumn{2}{c|}{\textbf{Modality}} &
  \multicolumn{1}{c}{} &
  \multicolumn{1}{c|}{} &
  \multicolumn{1}{c}{} &
  \multicolumn{1}{c}{} &
  \multicolumn{1}{c}{} &
  \multicolumn{1}{c}{} &
  \multicolumn{1}{c}{} &
  \multicolumn{1}{c}{} \\ \cmidrule{2-3}
\multicolumn{1}{c|}{\multirow{-2}{*}{\textbf{Method}}} &
  \textbf{Video} &
  \textbf{Audio} &
  \multicolumn{1}{c}{\multirow{-2}{*}{\textbf{\begin{tabular}[c]{@{}c@{}}\#Pairs \\ (M)\end{tabular}}}} &
  \multicolumn{1}{c|}{\multirow{-2}{*}{\textbf{\begin{tabular}[c]{@{}c@{}}LLM \\ size\end{tabular}}}} &
  \multicolumn{1}{c}{\multirow{-2}{*}{\textbf{AVSD}}} &
  \multicolumn{1}{c}{\multirow{-2}{*}{\textbf{\begin{tabular}[c]{@{}c@{}}MUSIC- \\ QA\end{tabular}}}} &
  \multicolumn{1}{c}{\multirow{-2}{*}{\textbf{AVSSD}}} &
  \multicolumn{1}{c}{\multirow{-2}{*}{\textbf{\begin{tabular}[c]{@{}c@{}}MSVD- \\ QA\end{tabular}}}} &
  \multicolumn{1}{c}{\multirow{-2}{*}{\textbf{\begin{tabular}[c]{@{}c@{}}MSRVTT- \\ QA \end{tabular}}}} &
  \multicolumn{1}{c}{\multirow{-2}{*}{\textbf{\begin{tabular}[c]{@{}c@{}}ActivityNet- \\ QA \end{tabular}}}} \\ \midrule[1pt]
Valley             & \tmark & \xmark  & 1.5   & 13B & -    & -    & -    & 65.4 & 45.7 & 26.5 \\
VideoChat         & \tmark & \xmark  & 25.0  & 7B  & -    & -    & -    & 56.3 & 45.0 & 26.5 \\
Video-ChatGPT     & \tmark & \xmark  & 0.9   & 7B  & -    & -    & -    & 64.9 & 49.3 & 35.2 \\
VTimeLLM         & \tmark & \xmark  & 0.7   & 7B  & -    & -    & -    & 69.8 & 58.8 & 45.5 \\
PandaGPT       & \tmark & \tmark & 128.0 & 13B & 26.1 & 33.7 & 32.7 & 46.7 & 23.7 & 11.2 \\
Macaw--LLM         & \tmark & \tmark & 0.3   & 13B & 34.3 & 31.8 & 36.1 & 42.1 & 25.5 & 14.5 \\
AV--LLM           & \tmark & \tmark & 1.6   & 7B  & 52.6 & 45.2 &     - & 67.3 & 53.7 & 47.2 \\
Video--LLaMA       & \tmark & \tmark & 2.8   & 13B & 36.7 & 36.6 & 36.7 & 51.6 & 29.6 & 12.4 \\
AVicuna           & \tmark & \tmark & 1.1   & 7B  & \underline{53.1} & 49.6 & -    & \underline{70.2} & \underline{59.7} & \underline{53.0} \\
Video-LLaMA2  & \tmark & \tmark & 2.0   & 7B  & $50.6^{*}$ & $\underline{66.3}^{*}$ & \textbf{71.4} & -    & -    & -    \\ \midrule[1pt]
\rowcolor[HTML]{DAE8FC} 
{\color[HTML]{333333} \ourapproach} &
  {\color[HTML]{333333} \tmark} &
  {\color[HTML]{333333} \tmark} & 
  {\color[HTML]{333333} 0.8} &
  {\color[HTML]{333333} 7B} & \fres{\textbf{55.1}}{+3.6}{ForestGreen}
  {\color[HTML]{333333} } & \fres{\textbf{69.8}}{+5.0}{ForestGreen}
  {\color[HTML]{333333} } & \fres{\underline{70.2}}{-1.7}{OrangeRed}
  {\color[HTML]{333333} } & \fres{\textbf{73.3}}{+4.2}{ForestGreen}
  {\color[HTML]{333333} } & \fres{\textbf{63.1}}{+5.4}{ForestGreen}
  {\color[HTML]{333333} } & \fres{\textbf{57.6}}{+8.0}{ForestGreen} \\ \bottomrule[2pt]
\end{tabular}
}
\vspace{-1em}
\end{table*}

\section{Experimental Evaluation of \ourapproach}
\label{expermient}
\label{dataset protocol}
\parlabel{Training Datasets}
\ourapproach is pretrained (Stage~\Romannum{1}) on 13.1M weakly aligned modality--text pairs (e.g., InternVid-10M, WavCaps, SensorCaps), and fine-tuned (Stages~\Romannum{2}--\Romannum{3}) on 510K high-quality QA pairs from \dname. See Appendix~\ref{supp_training_dataset} for details.


\parlabel{Validation Datasets}
We evaluate on seven audio-visual QA benchmarks spanning exocentric and egocentric domains: AVSD~\cite{avsd}, MUSIC-QA~\cite{musicavqa}, AVSSD~\cite{avssd}, MSVD-QA~\cite{avsd}, MSRVTT-QA~\cite{msrvttqa}, ActivityNet-QA~\cite{activitynetqa}, and EgoThink~\cite{egothink}, plus the 58K held-out test set from \dname (Appendix~\ref{supp:val_dataset}).
Evaluation metrics (GPT based) follow prior work~\cite{videochatgpt} as detailed in Appendix~\ref{supp:eval_metric}.


\parlabel{Baseline Models}
We compare against SOTA models across both domains. For egocentric QA: Valley~\cite{valley}, VideoChat~\cite{videochat}, VTimeLLM~\cite{vtimellm}, PandaGPT~\cite{pandagpt}, MacawLLM~\cite{macawllm}, AV-LLM~\cite{avllm}, Video-LLaMA~\cite{videollama1}, AVicuna~\cite{avicuna}, and Video-LLaMA2~\cite{videollama2}; for exocentric QA: OpenFlamingo~\cite{openflamingo}, BLIP-2.6~\cite{blip2}, VideoChat-7B~\cite{videochat}, LLaVA-1.5~\cite{llava15}, MiniGPT4~\cite{minigpt4}, InstructBLIP~\cite{instructblip}, LLaMA-Adapter~\cite{llamaadapter}, VideoLLaVA~\cite{videollava}, and ShareGPT4V~\cite{sharegpt4v}. All baselines use official checkpoints (See Appendix~\ref{supp_baselines}).

\begin{table*}[!htb]
\centering
\caption{Comparison of \ourapproach with MLLMs on the EgoThink (Reasoning) and AVS-QA benchmarks. \ourapproach outperforms across metrics and excels in reasoning. \textbf{Bold} and \underline{underline} indicate the best and second-best~scores.}
\vspace{-0.75em}
\label{tab_ego_av_main}
\resizebox{\textwidth}{!}{
\begin{tabular}{l|cccccccc}
\toprule[2pt]
\multicolumn{1}{c|}{}                  & \multicolumn{4}{c|}{EgoThink (Reasoning)}       & \multicolumn{4}{c}{\dname}  \\ \cmidrule(l){2-9} 
\multicolumn{1}{c|}{\multirow{-2}{*}{\textbf{Method}}} &
  \multicolumn{1}{c}{\textbf{Count}} &
  \multicolumn{1}{c}{\textbf{Compar}} &
  \multicolumn{1}{c}{\textbf{Situated}} &
  \multicolumn{1}{c|}{\textbf{Avg}} &
  \multicolumn{1}{c}{\textbf{Comp.}} &
  \multicolumn{1}{c}{\textbf{Coher.}} &
  \multicolumn{1}{c}{\textbf{Acc.}} &
  \multicolumn{1}{c}{\textbf{Avg}} \\ \midrule[1pt]
\multicolumn{1}{l|}{OpenFlamingo }  & 0.21 & 0.40 & 0.21 & \multicolumn{1}{c|}{0.27} & 0.31 & 0.34 & 0.27 & 0.31 \\
\multicolumn{1}{l|}{BLIP-2.6}      & 0.03  & 0.21 & 0.33 & \multicolumn{1}{c|}{0.19} & 0.22 & 0.26 & 0.21 & 0.23 \\
\multicolumn{1}{l|}{VideoChat}     & 0.36 & 0.39 & 0.32 & \multicolumn{1}{c|}{0.36} & 0.29 & 0.33 & 0.37 & 0.33 \\
\multicolumn{1}{l|}{LLaVA-1.5}     & 0.20 & 0.47 & 0.37 & \multicolumn{1}{c|}{34.7} & 0.46 & 0.47 & 0.52 & 0.48 \\
\multicolumn{1}{l|}{MiniGPT-4}     & 0.14 & \underline{0.48} & 0.31 & \multicolumn{1}{c|}{0.31} & 0.19 & 0.29 & 0.34 & 0.27 \\
\multicolumn{1}{l|}{InstructBLIP}  & 0.18 & 0.43 & \textbf{0.67} & \multicolumn{1}{c|}{0.42} & 0.33 & 0.37 & 0.35 & 0.35 \\
\multicolumn{1}{l|}{LLaMA-Adapter} & 0.29 & 0.39 & 0.25 & \multicolumn{1}{c|}{0.31} & 0.25 & 0.31 & 0.29 & 0.28 \\
\multicolumn{1}{l|}{PandaGPT}      & 0.19 & 0.52 & 0.53 & \multicolumn{1}{c|}{0.41} & 0.38 & 0.42 & 0.41 & 0.40 \\
\multicolumn{1}{l|}{VideoLLaVA}    & \underline{0.39} & 0.38 & 0.60 & \multicolumn{1}{c|}{\underline{0.46}} & 0.42 & 0.46 & 0.45 & 0.44 \\
\multicolumn{1}{l|}{ShareGPT4V}    & 0.30 & 0.38 & 0.66 & \multicolumn{1}{c|}{0.45} & \underline{0.64} & \underline{0.63} & \underline{0.59} & \underline{0.62} \\ \midrule[1pt]
\rowcolor[HTML]{C9DAF8} 
\multicolumn{1}{l|}{\cellcolor[HTML]{C9DAF8} \ourapproach} &
  \fres{\textbf{0.40}}{+2.7}{ForestGreen} &
  \fres{\textbf{0.54}}{+3.4}{ForestGreen} &
  \fres{\underline{0.66}}{-1.5}{OrangeRed} &
  \multicolumn{1}{c|}{\cellcolor[HTML]{C9DAF8}\fres{\textbf{0.54}}{+14.8}{ForestGreen}} &
  \fres{\textbf{0.71}}{+9.8}{ForestGreen} &
  \fres{\textbf{0.69}}{+8.7}{ForestGreen} &
  \fres{\textbf{0.61}}{+3.28}{ForestGreen} &
  \fres{\textbf{0.67}}{+7.5}{ForestGreen} \\ \bottomrule[2pt]
\end{tabular}
}
\vspace{-0.5em}
\end{table*}
\subsection{Quantitative Results}
\parlabel{Exocentric Audio-Visual}
Table~\ref{tab_av_main_result_exo} shows that \ourapproach outperforms SOTA models on video QA (by up to \textbf{8.0\%}) and AVQA (by \textbf{5.0\%}), surpassing QA-specific fusion models (e.g., AV-LLM, Macaw–LLM). These gains stem from \amodule's fine-grained, query-conditioned relevance scores, which enhance alignment and suppress irrelevant inputs. Performance is competitive but not superior on curated benchmarks like AVSSD, where modality-based relevance scoring may be less impactful due to limited cross-modal variability.

\parlabel{Egocentric Audio-Visual Results} 
Table~\ref{tab_ego_av_main} reports results on EgoThink and AVS-QA. \ourapproach achieves the highest overall performance--\textbf{53.5} average on EgoThink (+\textbf{14.6\%}) and \textbf{0.67} on AVS-QA (+\textbf{7.5\%})--with strong gains in Completeness (\textbf{0.71}, +\textbf{9.8\%}) and Correctness (\textbf{0.69}, +\textbf{8.7\%}). While baselines like OpenFlamingo-7B and BLIP-2.6-7B perform moderately (e.g., 21.0 on Count, 0.31 on Completeness), and VideoLLaVA-7B excels in specific categories (e.g., 66.0 in Situated), \ourapproach delivers the best overall scores.
\begin{figure*}[!htb]
\begin{minipage}{0.4\textwidth}
\centering
\captionsetup{type=table}
\caption{\dname results comparing \ourapproach with SOTA models using different modality combinations.}
\vspace{-0.5em}
\label{tab_main_res_avs}
\centering
\resizebox{\linewidth}{!}{
\begin{tabular}{l|ccc|cccc}
\toprule[2pt]
\multirow{-1}{*}{\textbf{Method}} & V & A & S & \multirow{-1}{*}{\textbf{Comp.}} & \multirow{-1}{*}{\textbf{Coher.}} & \multirow{-1}{*}{\textbf{Acc.}} & \multirow{-1}{*}{\textbf{Avg}} \\ 
\midrule[1.5pt]
\multicolumn{1}{l|}{}                                               & \tmark & \xmark  & \multicolumn{1}{c|}{\xmark}                         & 0.27 & 0.32 & 0.23 & 0.27 \\
\multicolumn{1}{l|}{\multirow{-2}{*}{Macaw-LLM}}                    & \tmark & \tmark & \multicolumn{1}{c|}{\xmark}                         & 0.38 & 0.46 & 0.34 & 0.39 \\ 
 \arrayrulecolor{shadecolor} \cmidrule[1pt](l){1-8}\arrayrulecolor{black}
\multicolumn{1}{l|}{}                                               & \tmark & \xmark  & \multicolumn{1}{c|}{\xmark}                         & 0.36 & 0.42 & 0.33 & 0.37 \\
\multicolumn{1}{l|}{\multirow{-2}{*}{Panda-GPT}}                    & \tmark & \tmark & \multicolumn{1}{c|}{\xmark}                         & 0.43 & 0.49 & 0.38 & 0.43 \\ 
\arrayrulecolor{shadecolor} \cmidrule[1pt](l){1-8}\arrayrulecolor{black}
\multicolumn{1}{l|}{}                                               & \tmark & \xmark  & \multicolumn{1}{c|}{\xmark}                         & 0.37 & 0.33 & 0.28 & 0.33 \\
\multicolumn{1}{l|}{\multirow{-2}{*}{VideoLLaMA}}                   & \tmark & \tmark & \multicolumn{1}{c|}{\xmark}                         & 0.48 & 0.51 & 0.41 & 0.47 \\
\arrayrulecolor{shadecolor} \cmidrule[1pt](l){1-8}\arrayrulecolor{black}
\multicolumn{1}{l|}{}                                               & \tmark & \xmark  & \multicolumn{1}{c|}{\xmark}                         & 0.51 & 0.54 & 0.43 & 0.49 \\
\multicolumn{1}{l|}{\multirow{-2}{*}{VideoLLaMA2}}                  & \tmark & \tmark & \multicolumn{1}{c|}{\xmark}                         & 0.56 & 0.59 & 0.51 & 0.55 \\ \midrule[1.5pt]
\rowcolor[HTML]{C9DAF8} 
\multicolumn{1}{l|}{\cellcolor[HTML]{C9DAF8}}                       & \tmark & \xmark  & \multicolumn{1}{c|}{\cellcolor[HTML]{C9DAF8}\xmark} & 0.61 & 0.62 & 0.46 & 0.56 \\
\rowcolor[HTML]{C9DAF8} 
\multicolumn{1}{l|}{\cellcolor[HTML]{C9DAF8}}                       & \tmark & \tmark & \multicolumn{1}{c|}{\cellcolor[HTML]{C9DAF8}\xmark} & \underline{0.71} & \underline{0.69} & \underline{0.61} & \underline{0.67} \\
\rowcolor[HTML]{C9DAF8} 
\multicolumn{1}{l|}{\multirow{-3}{*}{\cellcolor[HTML]{C9DAF8} \ourapproach}} & \tmark & \tmark & \tmark                                             & 
\textbf{0.78}
\ & 
\textbf{0.82}
\ & 
\textbf{0.73}
\ & 
\textbf{0.78}
\ \\
\bottomrule[2pt]
\end{tabular}
\vspace{-0.25em}
}
\end{minipage}
\begin{minipage}{0.58\textwidth}
\centering
\captionsetup{type=table}

\caption{
 Comparison under cross-modal mismatch scenarios. \ourapproach with Stage \Romannum{3} fine-tuning consistently outperforms baseline methods across all evaluation metrics and benchmarks, demonstrating superior robustness to modality perturbations.}
\label{tab_cross_modal}
\resizebox{\linewidth}{!}{
\begin{tabular}{l|cccc|cccc}
\toprule[2pt]
 &
  \multicolumn{1}{c}{} &
  \multicolumn{1}{c}{} &
  \multicolumn{1}{c}{} &
  \multicolumn{1}{c|}{} &
  \multicolumn{4}{c}{\textbf{\dname}} \\ \cmidrule(l){6-9} 
\multirow{-2}{*}{\textbf{Method}} &
  \multicolumn{1}{c}{\multirow{-2}{*}{\textbf{AVSD}}} &
  \multicolumn{1}{c}{\multirow{-2}{*}{\textbf{\begin{tabular}[c]{@{}c@{}}MUSIC \\ QA \end{tabular}}}} &
  \multicolumn{1}{c}{\multirow{-2}{*}{\textbf{\begin{tabular}[c]{@{}c@{}}MSVD \\ QA\end{tabular}}}} &
  \multicolumn{1}{c|}{\multirow{-2}{*}{\textbf{\begin{tabular}[c]{@{}c@{}}Activity\\Net-QA \end{tabular}}}} &
  \multicolumn{1}{c}{\textbf{Comp.}} &
  \multicolumn{1}{c}{\textbf{Cohr.}} &
  \multicolumn{1}{c}{\textbf{Acc.}} &
  \multicolumn{1}{c}{\textbf{Avg.}} \\ \midrule[1.5pt]
PandaGPT  &
  12.2 &
  13.8 &
  21.8 &
  7.9 &
  0.23 &
  0.29 &
  0.26 &
  0.26 \\
Macaw-LLM &
 18.1 &
 14.5 &
 22.2 &
 10.6 &
 0.11 &
 0.21 &
 0.19 &
 0.17 \\
AV-LLM &
  24.7 &
  22.1 &
  49.8 &
  26.8 &
  - &
  - &
  - &
  - \\
Video-LLaMA &
 17.9 &
 24.6 &
 31.5 &
 25.3 &
 0.28 &
 0.39 &
 0.33 &
 0.33 \\
AVicuna &
  34.1 &
  31.3 &
  51.7 &
  31.9 &
  - &
  - &
  - &
  - \\
Video-LLaMA2 &
 43.2 &
 44.7 &
 52.1 &
 29.7 &
 0.51 &
 0.54 &
 0.48 &
 0.51 \\ \midrule[1.5pt]
\rowcolor[HTML]{C9DAF8} 
\frestwo{\ourapproach}{\Romannum{1}, \Romannum{2}}
{black} &
  \underline{51.9} &
  \underline{63.7} &
  \underline{66.4} &
  \underline{52.6} &
  \underline{0.69} &
  \underline{0.71} &
  \underline{0.64} &
  \underline{0.68} \\
\rowcolor[HTML]{C9DAF8} 
\frestwo{\ourapproach}
{\Romannum{1} -- \Romannum{3}}
{black} 
&
  \textbf{54.9} &
  \textbf{69.2} &
  \textbf{72.8} &
  \textbf{57.2} &
  \textbf{0.76} &
  \textbf{0.79} &
  \textbf{0.71} &
  \textbf{0.75} \\ 
  \bottomrule[2pt]
\end{tabular}
}
\end{minipage}
\vspace{-0.5em}
\end{figure*}



\parlabel{Sensor-Aware Evaluation on \dname}
Table~\ref{tab_main_res_avs} reports results on \dname across modalities (V/A/S) and metrics (Completeness, Coherence, Accuracy, Avg). \ourapproach performs better than baselines like VideoLLaMA2 with A+V fusion (+21.8\% avg). However, \ourapproach with A+V+S achieves an additional performance gain of 16.4\% -- highlighting the benefit of sensor modality and sensor-aware reasoning. These results validate the importance of query-guided sensor integration for context-rich QA.

\parlabel{Cross-modal mismatch} 
Table~\ref{tab_cross_modal} shows \ourapproach effectively handles cross-modal mismatch. Trained with Stages~\textbf{\Romannum{1}} and \textbf{\Romannum{2}}, it outperforms prior SOTA on AVQA by 30--79\%. On \dname, Stage~\textbf{\Romannum{3}} fine-tuning boosts performance to 0.71--0.79, surpassing Video-LLaMA2 (0.51--0.54). These gains stem from \amodule’s query-to-token alignment, which emphasizes semantically relevant tokens even under modality misalignment.

\begin{figure*}[!htb]
\begin{minipage}{0.65\textwidth}
\centering
\captionsetup{type=table}
\caption{Ablation on \textbf{training stages} (\Romannum{2} \& \Romannum{3}), conditioning $\mathcal{L}_{\amodule}$ on $\mathbf{Z}$ ($\mathcal{L}_{\amodule}|\mathbf{Z}$) vs. $\mathbf{C}$ ($\mathcal{L}_{\amodule}|\mathbf{C}$), and regularization strength $\boldsymbol{\lambda}$.}
\label{tab_ablation_main}
\resizebox{\linewidth}{!}{
\begin{tabular}{l|c|c|ccccc|cccc}
\toprule[2pt]
\multirow{2}{*}{\begin{tabular}[c]{@{}c@{}}\textbf{Training} \\ \textbf{Stage}\end{tabular}} &
  \multicolumn{1}{c|}{\multirow{2}{*}{\textbf{Loss}}} &
  \multirow{2}{*}{$\boldsymbol{\lambda}$} &
  \multirow{2}{*}{\textbf{AVSD}} &
  \multirow{2}{*}{\textbf{\begin{tabular}[c]{@{}c@{}}MUSIC \\ QA \end{tabular}}} &
  \multirow{2}{*}{\textbf{AVSSD}} &
  \multirow{2}{*}{\textbf{\begin{tabular}[c]{@{}c@{}}MSVD \\ QA \end{tabular}}} &
  \multirow{2}{*}{\textbf{\begin{tabular}[c]{@{}c@{}}Activity\\Net-QA \end{tabular}}} &
  \multicolumn{4}{c}{\textbf{\dname}} \\ \cmidrule(l){9-12} 
                        & \multicolumn{1}{c|}{}     &       &      &      &      &      &      & \textbf{Comp.} & \textbf{Cohr.} & \textbf{Acc.} & \textbf{Avg.} \\ \midrule[1.5pt]
\multirow{2}{*}{\begin{tabular}[c]{@{}c@{}}\textbf{Up to} \\ \textbf{Stage \Romannum{2}}\end{tabular}} & \begin{tabular}[c]{@{}c@{}}$\boldsymbol{\mathcal{L}_{\amodule}|\mathbf{Z}}$ 
\end{tabular}                         & -     & 45.2 & 53.2 & 58.8 & 60.3 & 45.1 & 0.38           & 0.52           & 0.42          & 0.44          \\
  \arrayrulecolor{shadecolor} \cmidrule[1pt](l){2-12}\arrayrulecolor{black}
                        &      \begin{tabular}[c]{@{}c@{}}$\boldsymbol{\mathcal{L}_{\amodule}|\mathbf{C}}$ 
                        \end{tabular}                     & -     & 48.7 & 57.7 & 61.5 & 63.9 & 51.2 & 0.42           & 0.57           & 0.47          & 0.49          \\ \midrule[1.5pt]
\multirow{5}{*}{\begin{tabular}[c]{@{}c@{}}\textbf{Up to} \\ \textbf{Stage \Romannum{3}}\end{tabular}} & \begin{tabular}[c]{@{}c@{}}w/o  $\boldsymbol{\mathcal{L}_{reg}}$\end{tabular}                 & -     & 40.7 & 48.5 & 59.3 & 61.5 & 43.2 & 0.29           & 0.41           & 0.34          & 0.35          \\ \cmidrule(l){2-12} 
                        & \multirow{4}{*}{\begin{tabular}[c]{@{}c@{}}with \\ $\boldsymbol{\mathcal{L}_{reg}}$\end{tabular}} & 1     & 41.5 & 45.3 & 53.2 & 57.9 & 39.7 & 0.23           & 0.37           & 0.29          & 0.30          \\ 
                        &                           & 0.1   & 48.3 & 56.2 & 54.7 & 64.2 & 45.8 & 0.62           & 0.69           & 0.59          & 0.63          \\
                        &                           & 0.01  & 52.2 & 61.8 & 61.2 & 68.1 & 51.6 & 0.71           & 0.78           & 0.68          & 0.72          \\
                        &                           & 0.001 & \textbf{55.1} & \textbf{69.8} & \textbf{70.2} & \textbf{73.3} & \textbf{57.6} & \textbf{0.78}           & \textbf{0.82}           & \textbf{0.73}          & \textbf{0.78}          \\ \bottomrule[2pt]
\end{tabular}
}
\end{minipage}
\begin{minipage}{0.34\textwidth}
\centering
\captionsetup{type=table}
\caption{Effect of $\mathbf{W}^R$. \amodule outperforms with fewer parameters.}
\vspace{-1em}
\label{tab:wr}
\resizebox{\textwidth}{!}{
\begin{tabular}{c|ccc
>{\columncolor[HTML]{C9DAF8}}c}
\toprule[2pt]
\multicolumn{1}{l|}{\textbf{Method}} &
  \multicolumn{1}{l}{\textbf{\begin{tabular}[c]{@{}c@{}}Raw \\ attention\end{tabular}}} &
  \multicolumn{1}{l}{\textbf{\begin{tabular}[c]{@{}c@{}}Q - \\ Former\end{tabular}}} &
  \multicolumn{1}{l}{\textbf{HierarQ}} &
  \multicolumn{1}{l}{\cellcolor[HTML]{C9DAF8}\amodule} \\ \midrule[1.5pt]
\textbf{\#Params $\boldsymbol{\downarrow}$}                                                   & \textbf{41M}  & 188M & 390M & 45M  \\
\arrayrulecolor{shadecolor} \cmidrule[1pt](l){1-5}\arrayrulecolor{black}
\textbf{AVSD}                                                      & 29.1 & 36.7 & -    & \textbf{55.1} \\
\arrayrulecolor{shadecolor} \cmidrule[1pt](l){1-5}\arrayrulecolor{black}
\textbf{\begin{tabular}[c]{@{}c@{}}MUSIC-QA\end{tabular}}       & 23.6 & 36.6 & -    & \textbf{69.8} \\
\arrayrulecolor{shadecolor} \cmidrule[1pt](l){1-5}\arrayrulecolor{black}
\textbf{\begin{tabular}[c]{@{}c@{}}MSVD-QA\end{tabular}}        & 42.2 & 51.6 & 66.2 & \textbf{73.3} \\
\arrayrulecolor{shadecolor} \cmidrule[1pt](l){1-5}\arrayrulecolor{black}
\textbf{\begin{tabular}[c]{@{}c@{}}ActivityNet\\-QA\end{tabular}} & 12.1 & 12.4 & 57.2 & \textbf{57.6} \\
\arrayrulecolor{shadecolor} \cmidrule[1pt](l){1-5}\arrayrulecolor{black}
\textbf{\begin{tabular}[c]{@{}c@{}}MSRVTT\\ -QA\end{tabular}}      & 23.1 & 29.6 & 54.1 & \textbf{63.1} \\ \bottomrule[2pt]
\end{tabular}
}
\end{minipage}
\vspace{-1em}
\end{figure*}
\subsection{Ablation Study}
\label{ablation_study}
\parlabel{Training Stages and Loss Conditioning}
We ablate training stages, loss formulation, and regularization strength across six QA benchmarks (Table~\ref{tab_ablation_main}). Conditioning $\mathcal{L}_{\amodule}$ on contextual embeddings $\mathbf{C}$ (vs. raw $\mathbf{Z}$) in Stage~\Romannum{2} improves performance (e.g., \dname Avg: 0.49 vs. 0.44), confirming the value of context in alignment. Adding regularization in Stage~\Romannum{3} boosts robustness but is sensitive to $\lambda$: a high value (1.0) hurts performance (\dname Avg: 0.30), while $\lambda = 0.001$ yields the best results--raising AVS-QA Avg to 0.78 (+43\%), Coherence to 0.82 (+15.9\%), and Accuracy to 0.73 (+16.4\%). Similar gains appear on ActivityNet-QA (+18.4\%) and MUSIC-QA (+24.5\%). Overall, best performance is achieved with Stage~\Romannum{3}, context-aware $\mathcal{L}_{\amodule}$, and $\lambda = 0.001$--highlighting the synergy between structured alignment and calibrated regularization.

\parlabel{Effect of Learnable Relevance Projection ($\mathbf{W}^R$)} 
Table~\ref{tab:wr} compares \amodule’s learnable projection head $\mathbf{W}^R$  against raw attention and two state-of-the-art token relevance methods: Q-Former~\cite{blip2} and HierarQ~\cite{azad2025hierarq}. \amodule achieves the highest accuracy across all benchmarks while using fewer parameters (45M vs. 188M/390M). By transforming attention scores into query-conditioned relevance weights, $\mathbf{W}^R$ enables efficient and interpretable cross-modal alignment.
Additional ablations -- including encoder choices, LoRA rank, token selection -- are provided in Appendix~\ref{supp_ablation_study}, along with qualitative examples in Appendix~\ref{supp_qualitative}.

\section{Conclusion}
In this paper, we present \ourapproach, a unified framework for multimodal question answering that integrates video, audio, and sensor inputs via query-aware alignment, enabling robust reasoning under modality disagreement. To support this, we release \dname--the first large-scale dataset of synchronized \{\texttt{Audio, Video, Sensor, QA}\} quadruples--curated via an automated actor-evaluator-critic pipeline. Spanning egocentric settings and four QA types, \dname enables comprehensive benchmarking. Our three-stage training--modality pretraining, query-conditioned alignment, and perturbation-aware fine-tuning--drives consistent gains across diverse multimodal QA benchmarks. These results underscore the importance of structured, query-aware reasoning in handling real-world modality mismatch.

\newpage
\section{Limitations}
 While \ourapproach provides a strong foundation for multimodal question answering over audio, video, and sensor inputs, our current experiments are limited to a single backbone model, \texttt{Qwen-Instruct-7B}, due to computational constraints. We do not explore larger LLM variants (e.g., 13B or 70B), which could further improve performance but require significantly more resources. Additionally, we leave the investigation of alternative language backbones and more advanced fusion strategies (e.g., retrieval-augmented alignment, memory-based conditioning) as future work.

 We also note that for longer recordings (exceeding $\sim$5 minutes), particularly those involving visually dense scenes, \ourapproach occasionally underperforms on vision-heavy queries. This is likely caused by our uniform frame selection strategy, which may miss critical visual cues in longer videos because of sparse temporal sampling. Incorporating adaptive or query-guided frame selection could mitigate this issue and improve temporal grounding.

 Finally, training \ourapproach is computationally expensive. Our current setup required approximately 120 hours on 4 NVIDIA A100 GPUs (each with 80 GB of memory). While the design is efficient at inference time due to early token filtering, future work could further reduce training cost through distillation or parameter sharing across modalities.

\parlabel{Future Directions} 
Future work on \ourapproach includes exploring joint training strategies across modalities to enable deeper cross-modal interactions and more robust representation learning. Incorporating a saliency-aware frame selection mechanism may further improve performance on long-form, visually complex inputs. Additionally, reducing or eliminating the need to fine-tune the LLM backbone when introducing new modalities remains an open challenge. Addressing this could significantly improve the scalability, adaptability, and deployment efficiency of multimodal language models.


 \section{Ethical Considerations}
 \label{ethical_statement}
 The \dname dataset is derived entirely from publicly released egocentric datasets (Ego4D \cite{ego4d} and EPIC-Kitchens \cite{epic-kitchen}) that include usage licenses permitting research redistribution. Our processing pipeline does not introduce new identity annotations, and we do not extract or distribute personally identifiable metadata. \dname contains synthetic question–answer pairs generated from visual, auditory, and sensor summaries, and no raw video, audio, or IMU recordings are included in the release. We follow best practices for anonymization and respect the original datasets’ ethical use guidelines.
 \section{Risk Statement}
 Our multimodal language model integrates audio, visual, and sensor inputs to enhance reasoning, but it raises several concerns. First, misuse of MLLMs in surveillance, biometric inference, or manipulation of multi-sensory content raises ethical concerns regarding user privacy and consent, especially when applied to egocentric or sensor-rich environments. Additionally, the interpretability of cross-modal reasoning remains limited, making it difficult to identify failure cases or mitigate hallucinations across modalities. We recommend careful deployment of such systems with human oversight, ongoing auditing of training data sources, and future work on explainability and robust alignment to reduce these risks.

\section*{Acknowledgment}
This research was supported by funding from the NSF CNS-2347692. Results in this paper were obtained in part using a high-performance computing system acquired through NSF MRI grant DMS-1337943 to WPI. We gratefully acknowledge their support in enabling this work.
\bibliography{reference}

\clearpage
\appendix
\section{More Related Works}
\label{supp_related_works}
This section includes additional models, datasets, and encoder variants relevant to our work that were not cited in the related work of the main paper due to space constraints. We list them here for completeness and to acknowledge recent progress in MLLMs and sensor-grounded QA.

\parlabel{Large Language Models}
Mixtral~\cite{mixtral}, Vicuna~\cite{vicuna}, Phi~\cite{phi}, OPT~\cite{opt}, PaLM~\cite{palm}


\parlabel{Sensor MLLMs}
MentalLLM \cite{mentalllm}, IMUGPT2.0 \cite{imugpt}, Sensor2Text \cite{sensor2text}, Penetrative AI \cite{xu2024penetrative}, PH-LLM~\cite{cosentino2024towards}, PHIA~\cite{merrill2024transforming}

\parlabel{Feature Alignment}
VLMo~\cite{bao2022vlmo}, FILIP~\cite{yao2021filip}, ALIGN~\cite{li2021align}, ImageBind~\cite{imagebind}, CoCa~\cite{yu2022coca}, EgoVLPv2~\cite{pramanick2023egovlpv2}, HiTeA~\cite{ye2023hitea}, Mixed Q-Former~\cite{wang2024omnivid}, Missingness resilient \cite{mohapatra24_interspeech, biswas2023locus}
\section{\dname Dataset Details}
\label{supp_dataset_curation}

\subsection{Curation and Statistical Summary}
\parlabel{Dataset Curation Stages}
In the Actor phase, we generated 387K question--answer pairs. The Evaluator filtered out 12.14\% based on predefined constraints. In the Critic phase, an additional 40K QA pairs were discarded based on aggregate scores from multiple critics. This results in a final dataset of 300K high-quality QA pairs used for training and evaluation.

\begin{figure}[!htb]
    \centering
    \includegraphics[width=\linewidth]{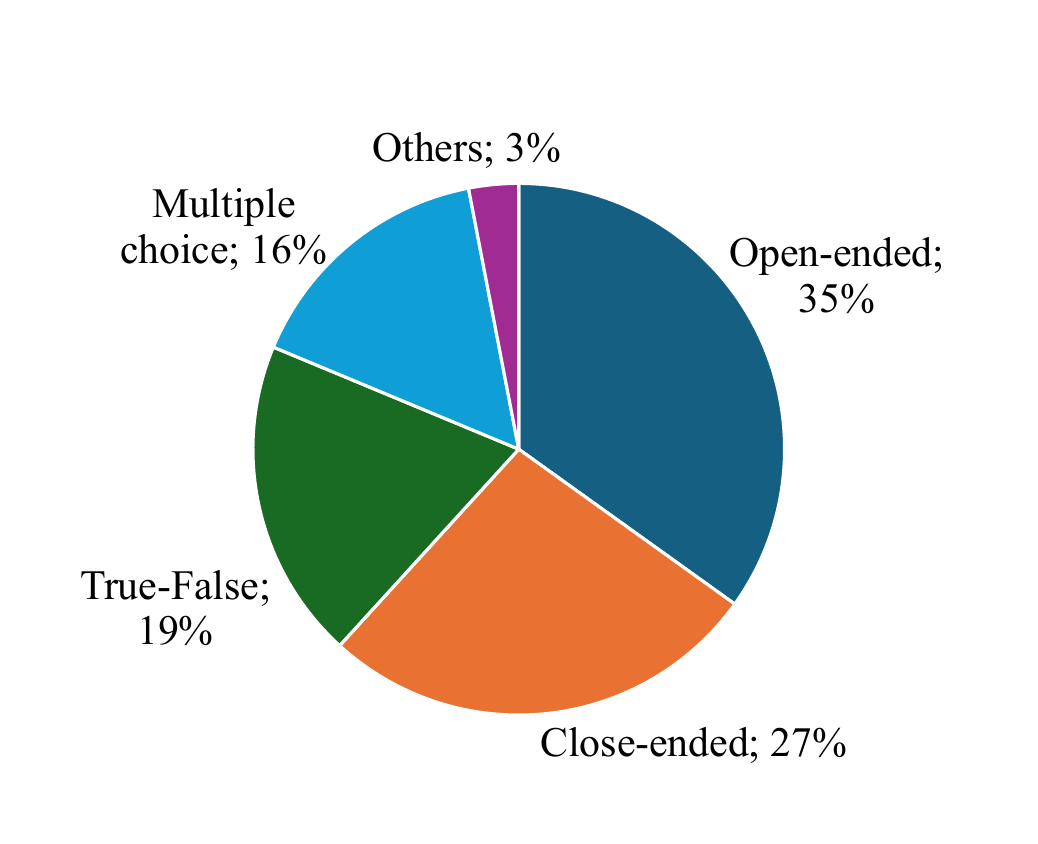}
    \caption{Distribution of question types in \dname. The dataset includes a diverse mix of open-ended, close-ended, true/false, multiple choice, and other formats, supporting comprehensive evaluation settings.}
    \label{fig:statistics}
\end{figure}
\parlabel{Distribution of Question Types}
\dname includes four primary question types to support diverse reasoning tasks: open-ended, close-ended, true/false, and multiple choice. Figure~\ref{fig:statistics} shows the distribution of these four categories. ``Others'' category include instructional or dialogue-style prompts that do not fit traditional QA formats. This variety enables comprehensive benchmarking across free-form generation and structured prediction settings.

\begin{figure}[!htb]
    \centering
    \includegraphics[width=\linewidth]{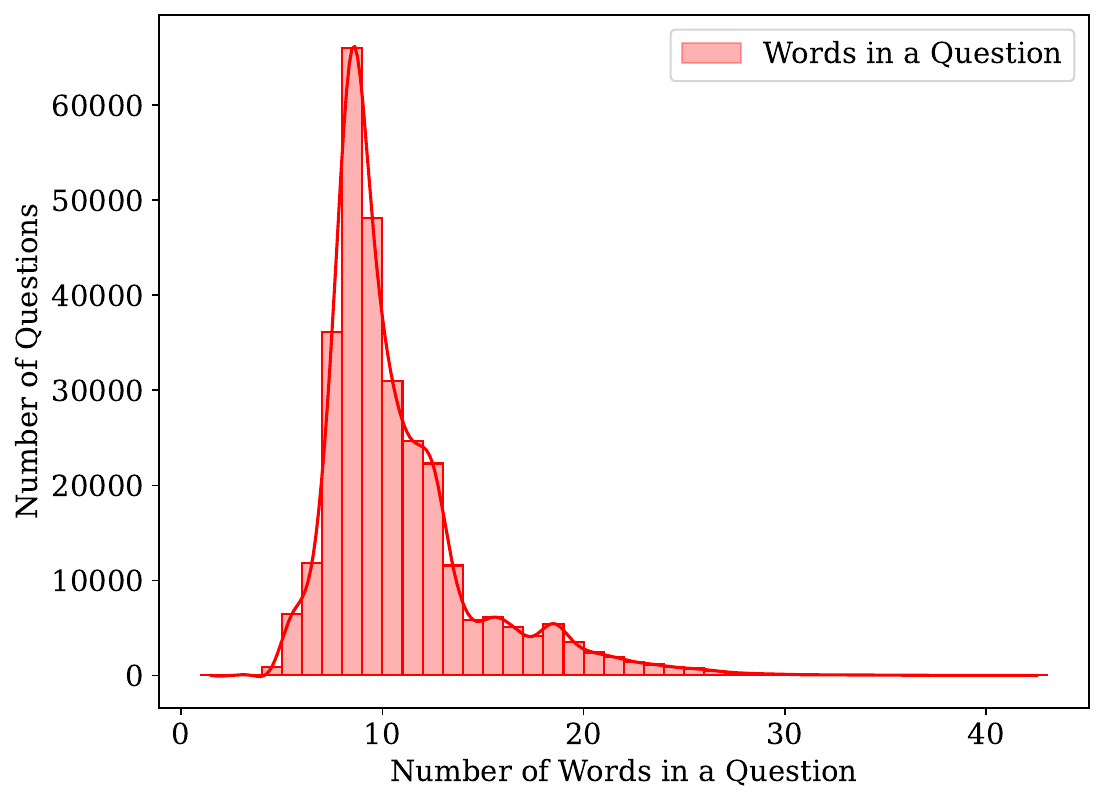}
    \caption{Length of questions has some variation due to different types of questions.}
    \label{fig:question_dist}
\end{figure}
\parlabel{Length Distribution of Questions and Answers}
We analyze the word-length distributions of questions and answers in \dname to better understand their linguistic diversity. As shown in Figure~\ref{fig:question_dist}, most questions are concise, with a mode around 9--10 words and a long-tail distribution extending up to 40 words. This variation arises from the presence of both short, structured formats (e.g., true/false, multiple choice) and more descriptive open-ended queries.

\begin{figure}[!htb]
    \centering
    \includegraphics[width=\linewidth]{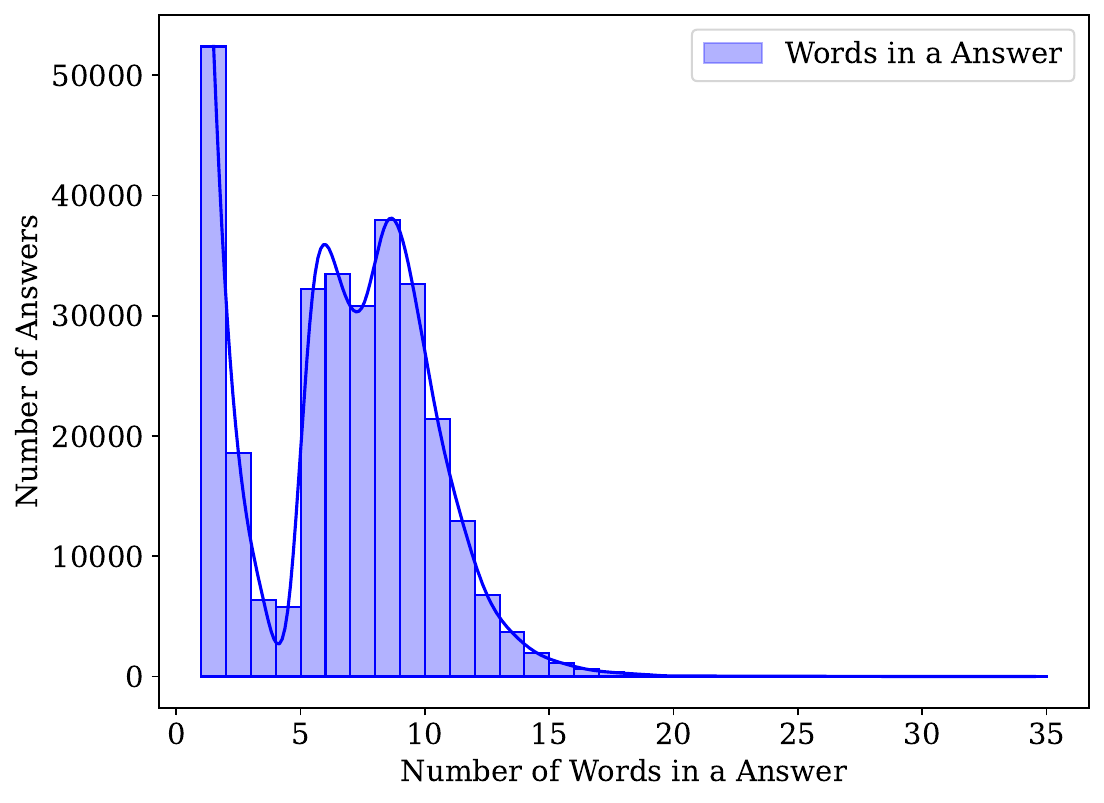}
    \caption{True/false and multiple choice questions often lead to one-word answers, while open-ended and close-ended formats yield a broader distribution of answer lengths.}
    \label{fig:answer_dist}
\end{figure}
Figure~\ref{fig:answer_dist} shows that a large number of answers consist of a single word, primarily due to true/false and multiple choice formats. In contrast, close-ended and open-ended questions yield longer and more varied responses, contributing to a broad distribution that peaks between 3--10 words and extends beyond 25 words. These distributions highlight the reasoning and generation challenges posed by \dname.

\parlabel{License}
\dname is released under a CC-BY 4.0 license, along with the full generation pipeline, including prompts, templates, and filtering scripts.

\subsection{Quality Ranking via LLM Scoring}
\label{supp_dataset_scorer}

To evaluate the quality of multi-modal (audio, video, sensor) question-answer pairs, we design a set of five quality assessment axes. Each axis is rated on a 5-point Likert scale (1 = poor, 5 = excellent) by large language models (LLMs) using structured prompts:

\parlabel{Answerability} 
Evaluates whether the question is answerable based on the provided multi-modal context. A high score indicates that the combined modalities contain sufficient and coherent information to support a correct and complete answer.

\parlabel{Hallucination Robustness} 
Measures the extent to which the answer avoids introducing information not grounded in the provided modalities. Higher scores indicate reliable adherence to the multi-modal context, while lower scores reflect a greater risk of hallucination.

\parlabel{Cross-Modal Grounding} 
Assesses the degree to which the answer integrates information across modalities (e.g., referencing audio to explain visual content). Higher scores reflect strong cross-modal coherence and accurate alignment with modality-specific cues relevant to the question.

\parlabel{Specificity} 
Measures the level of detail and precision in the answer relative to the question. Higher scores indicate clear, specific, and well-defined responses that avoid vague or generic statements, offering informative and actionable insights.

\parlabel{Relevance} 
Measures how directly the answer addresses the intent and scope of the question. Higher scores indicate focused, contextually appropriate responses that are clearly aligned with the queried scenario and available modalities.

Each QA pair is scored across the five axes by LLaVA-1.5\cite{llava15}, Gemeni Pro \cite{gemini}, Qwen-VL \cite{qwenvl}, GPT-4o \cite{gpt4}, LLaMA-3 \cite{llama3} in a zero-shot setting. We compute the final quality score by averaging the axis-level ratings.  We discard QA pairs where $\geq$2 axes receive a score $<$3 from at least 3 of 5 LLMs. This threshold was chosen based on alignment with human judgment (see Appendix~\ref{human_evaluation}).

\subsection{Human Evaluation}
\label{human_evaluation}
We conducted a human evaluation on a randomly selected subset of 300 question-answer pairs from \dname. Two expert annotators independently reviewed each sample and assigned quality ratings based on the accompanying video, audio, and sensor data. Ratings follow the same 5-point Likert format as the LLM scorer. 

We categorized the pairs based on human agreement:  
\textit{Satisfied} (both annotators rate $\geq$4),  
\textit{Okay} (mixed rating: one $\geq$4, one $<$4), and  
\textit{Not Satisfied} (both <4).  
We observe 81\% Satisfied, 7\% Okay, and 12\% Not Satisfied.

\textbf{This aligns closely with the filtering performed by our LLM critic, which rejected 40K of the initial 340K QA pairs (11.76\%), indicating strong agreement between human and automatic judgments.}
This suggests that our LLM-based scoring framework is a reliable proxy for human evaluation at scale.

We recruited two annotators through internal advertisements at the host institution. Both male annotators were between 25--35 years old and had a basic understanding of large language models. Participation was voluntary, and no financial incentives were provided.


\subsection{Prompt for Dataset Curation}
\begin{figure}[!htb]
    \centering
    \includegraphics[width=\linewidth]{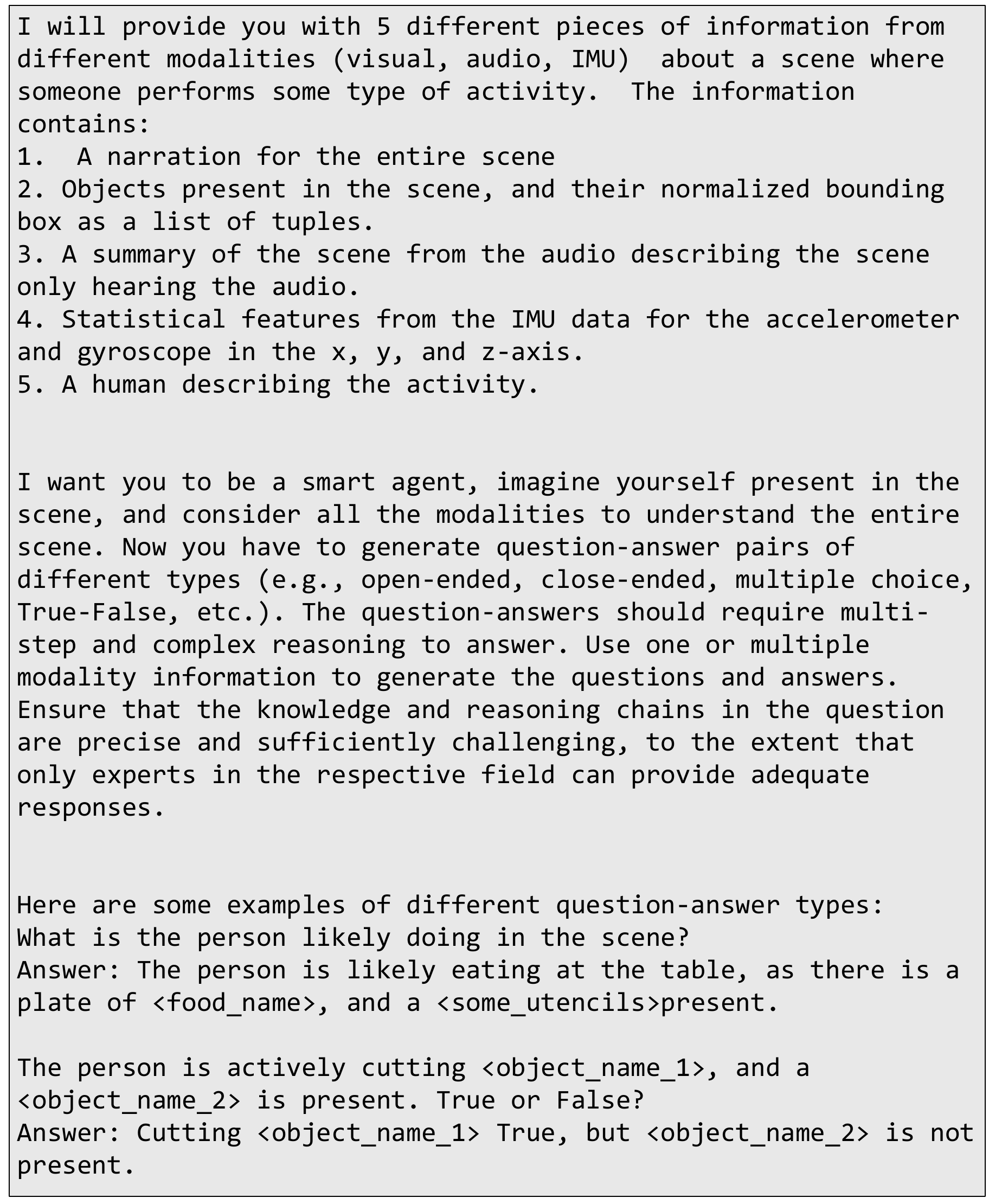}
    \caption{\textbf{System prompt used for generating questions and answers in Actor phase.}}
    \label{fig:actor_system}
\end{figure}

\begin{figure}[!htb]
    \centering
    \includegraphics[width=\linewidth]{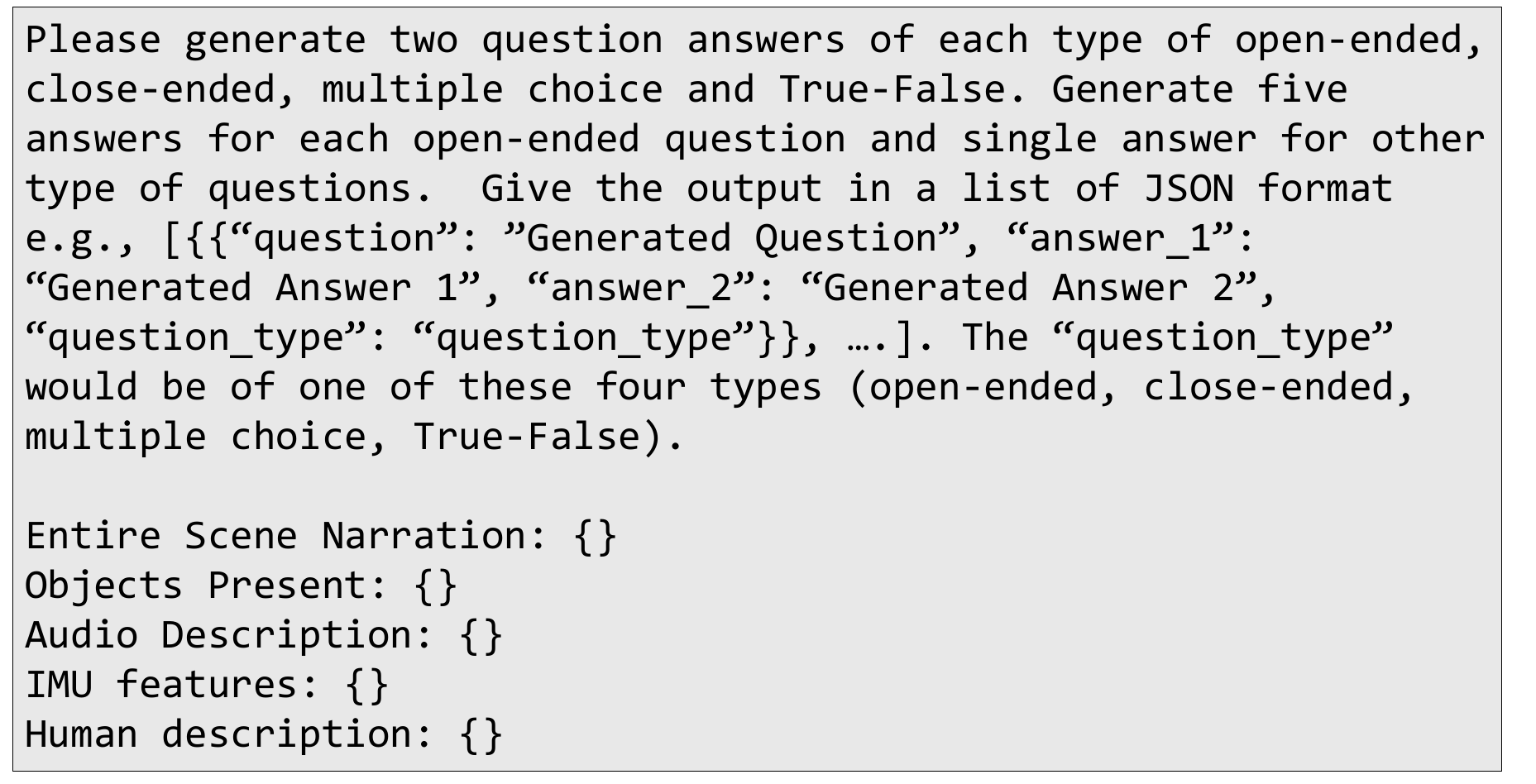}
    \caption{\textbf{User prompt used for generating questions and answers in Actor phase.}}
    \label{fig:actor_user}
\end{figure}

\begin{figure}[!htb]
    \centering
    \includegraphics[width=\linewidth]{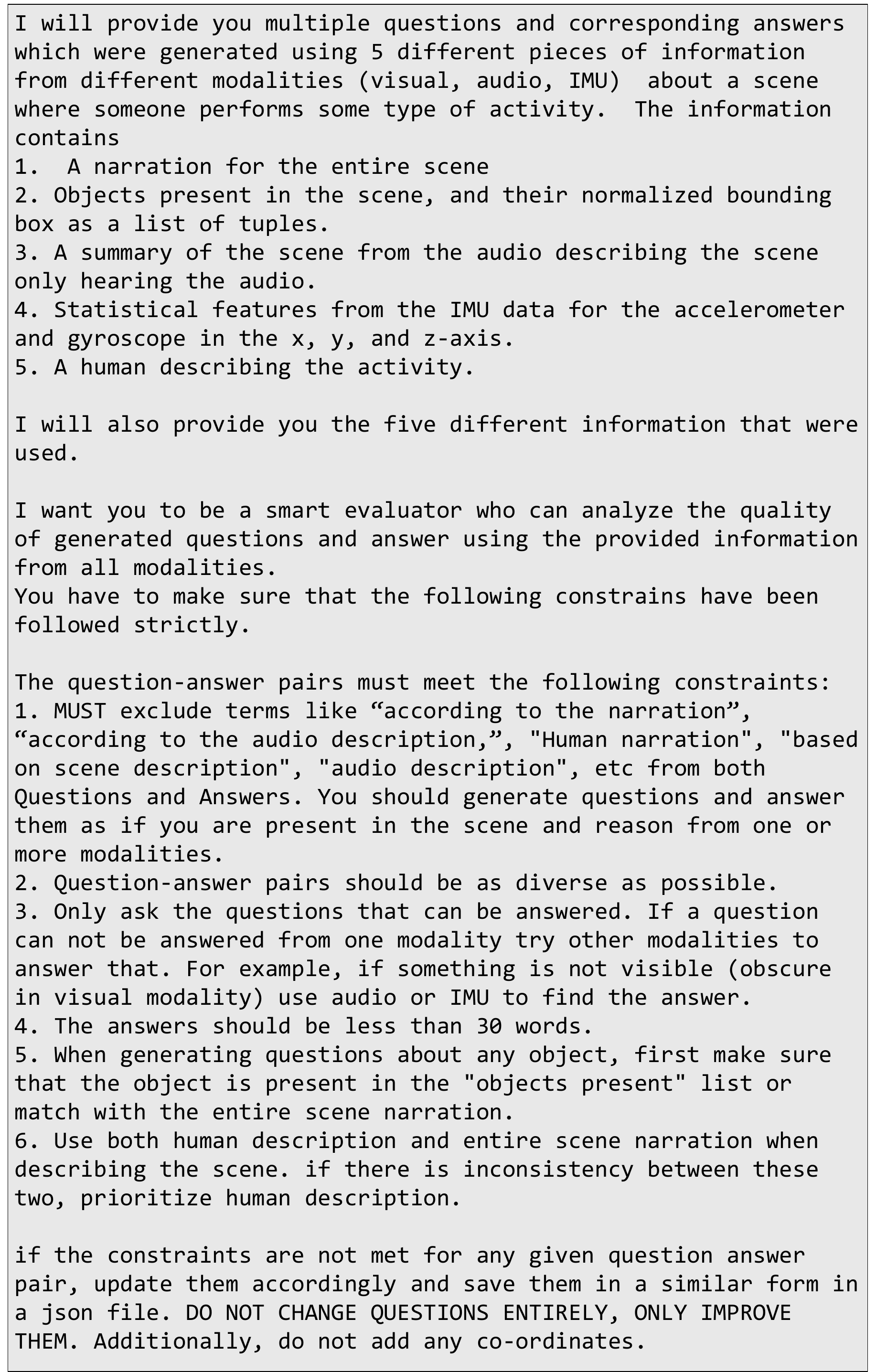}
    \caption{\textbf{System prompt used for generating questions and answers in Evaluator phase.} The constraints ensure avoiding some phrases or groups of words to enhance the quality of question-answer pairs.}
    \label{fig:evaluator_system}
\end{figure}

\begin{figure}[!htb]
    \centering
    \includegraphics[width=\linewidth]{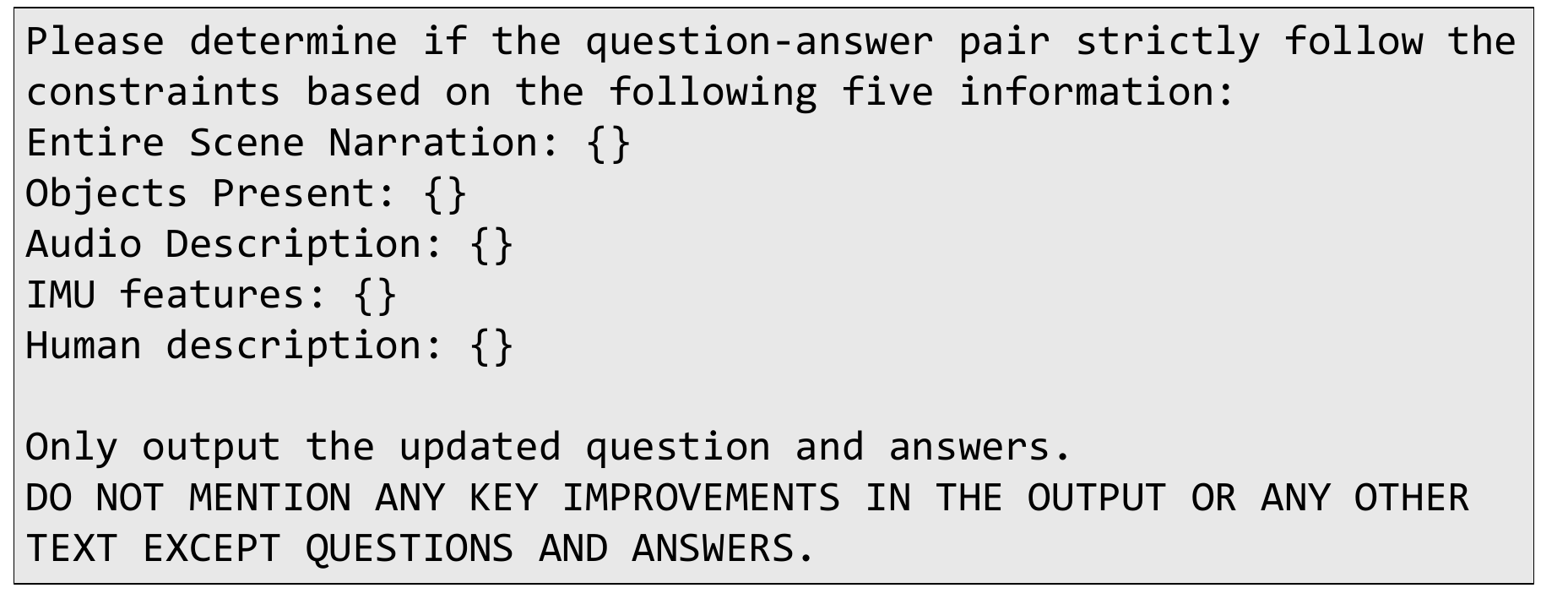}
    \caption{\textbf{User prompt used for generating questions and answers in Evaluator phase.}}
    \label{fig:evaluator_user}
\end{figure}

\begin{figure}[!htb]
    \centering
    \includegraphics[width=\linewidth]{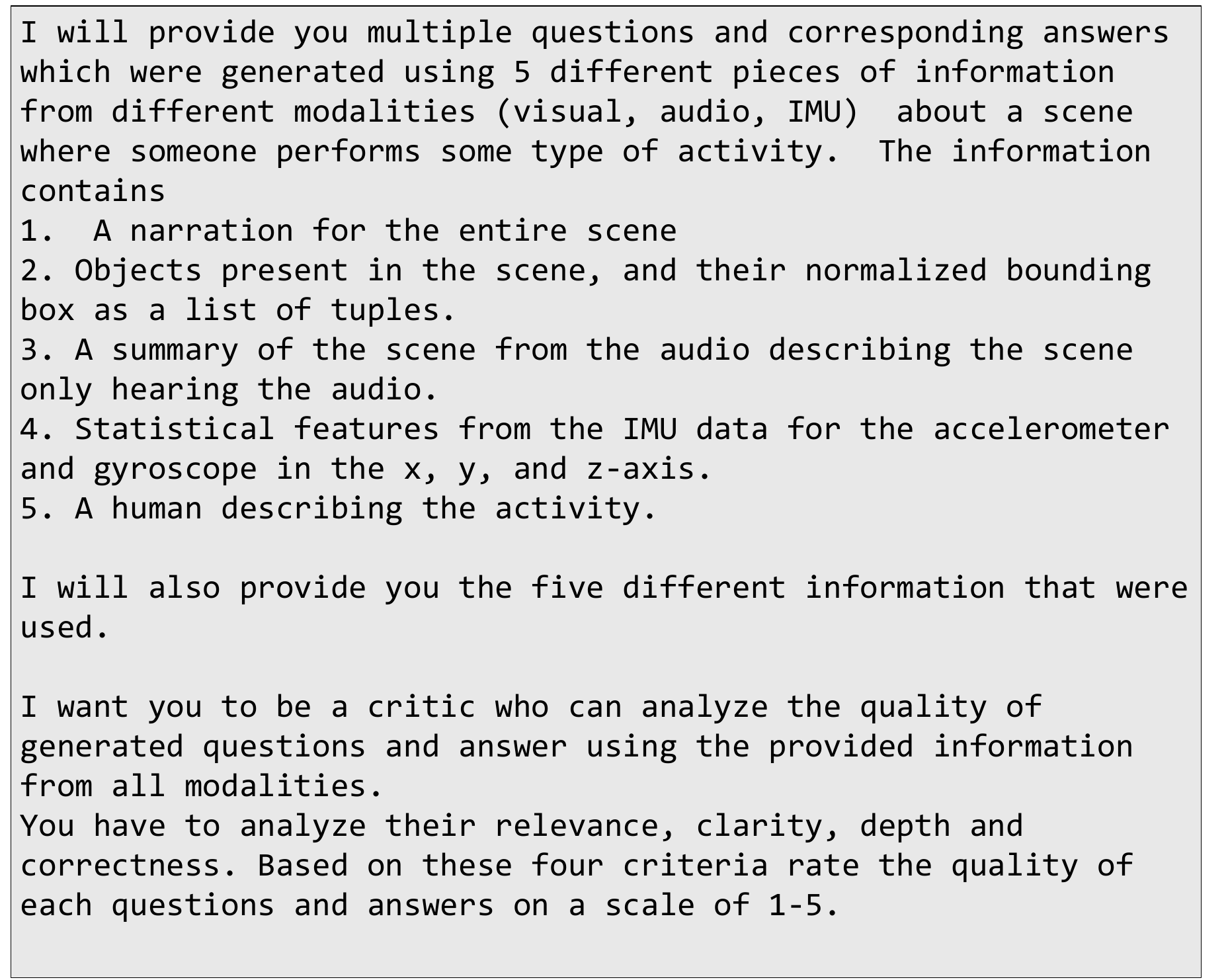}
    \caption{\textbf{System prompt used for generating questions and answers in Critic phase.}}
    \label{fig:critic_system}
\end{figure}

\begin{figure}[!htb]
    \centering
    \includegraphics[width=\linewidth]{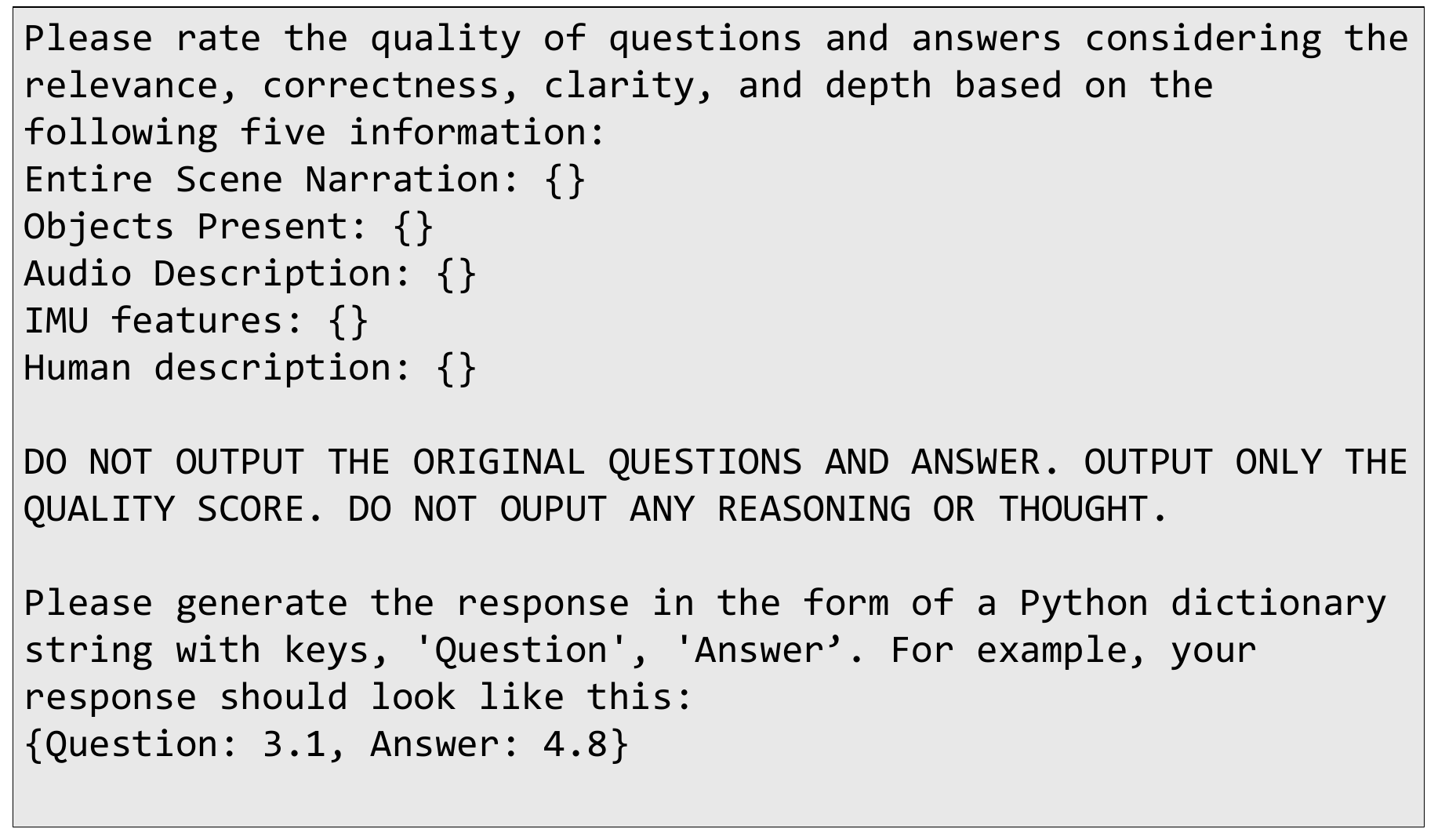}
    \caption{\textbf{User prompt used for generating questions and answers in Critic phase.}}
    \label{fig:critic_user}
\end{figure}

We use a structured Actor--Evaluator--Critic pipeline for automatic generation and refinement of question--answer pairs. Figures~\ref{fig:actor_system}--\ref{fig:critic_user} show the system and user prompts used at each stage of this pipeline.

In the \textbf{Actor phase}, a language model is provided with multimodal scene descriptions—including audio, video, IMU data summaries, and human narration—and is prompted to generate diverse questions spanning open-ended, close-ended, multiple choice, and true/false formats. The prompt encourages context-aware and modality-specific reasoning (see Figures~\ref{fig:actor_system}--\ref{fig:actor_user}).

In the \textbf{Evaluator phase}, a second model verifies the answerability, modality grounding, and factual correctness of each QA pair. The system prompt (Figure~\ref{fig:evaluator_system}) outlines constraints regarding modality coverage, object grounding, and language consistency. The human prompt (Figure~\ref{fig:evaluator_user}) ensures no hallucinated corrections are introduced—only local improvements to existing QA pairs.

In the Critic phase, large language models are prompted to rate the quality of each generated question--answer pair using four dimensions: relevance, correctness, clarity, and depth. As shown in Figures~\ref{fig:critic_system}--\ref{fig:critic_user}, the system prompt instructs the model to consider all five available modality-specific inputs (narration, object list, audio summary, IMU features, and human description) before assigning a score.

The user prompt standardizes the response format and explicitly prohibits speculative reasoning or textual justification—ensuring consistent, numerical evaluations across samples. Each QA pair receives two scores (one for the question, one for the answer), which are then aggregated across multiple critics to determine inclusion in the final dataset. QA pairs with low aggregate scores are discarded during the final curation step.

This prompt engineering strategy supports diverse and high-quality QA generation without human-in-the-loop authoring.


\section{Additional Model Architecture Details}
\label{supp_arch}

\subsection{LIMU-BERT Pre-Training}
\label{supp_limubert_pretraining}
As our sensor encoder, we employ LIMU-BERT~\cite{limubert}, a multi-head attention-based encoder-decoder architecture. LIMU-BERT is a lightweight, BERT-inspired self-supervised representation learning model designed for mobile IMU (Inertial Measurement Unit) sensing applications. It processes unlabeled IMU data—accelerometer, gyroscope, and magnetometer readings—to learn generalizable features. The architecture incorporates a normalization and sensor fusion layer, followed by a transformer encoder with cross-layer parameter sharing to reduce model size. It adopts a span-masking version of the Masked Language Modeling (MLM) task to learn both distributional and temporal patterns from the IMU sequences. We adopt the official LIMU-BERT implementation under the MIT license for research use.

\subsection{Unimodal Encoder Pre-Training}
We use the VideoLLaMA2~\cite{videollama2} codebase for pre-training the vision encoder. The encoder is initialized from a SigLIP checkpoint and fine-tuned with instructional video datasets included in the VideoLLaMA2 training suite. This setup enables the model to learn temporal and spatial reasoning over egocentric and exocentric scenes. The code is released under the Apache 2.0 license and used strictly for research purposes.

\subsection{Projection Layer}
\label{supp_projection_layer}
Each modality-specific encoder output is projected to the LLM input dimension using a tailored strategy.
The output of the audio encoder is projected through a two-layer multi-layer perceptron (MLP) to align with the LLM dimension. For the video encoder output, we use a spatio-temporal convolutional (STC) connector for spatio-temporal learning of the video. STC connector uses RegStage \cite{radosavovic2020designing} with 3D convolution for downsampling the video output. We use a publicly available adaptation of the STC-connector in our implementation \cite{videollama2} under the license of Apache 2.0 for research purposes only.

\section{Cross-Modal Mismatch Generation and Robustness Evaluation}
\label{supp_modal_dicrepency}
\begin{algorithm*}[t]
\caption{Algorithm for generating Cross-Modal Mismatch}
\label{alg_cross_modal_mismatch}
\begin{algorithmic}[1]
\Function{GenerateCrossModalMismatch}{$D = \{a, v, s\}$}
    \State Initialize $D' = \{a', v', s'\} \gets \{a, v, s\}$

    \State Define $P_\text{audio} \gets \{\textsc{AddNoise}, \textsc{Reverse}, \textsc{ReplaceWithIrrelevant}, \textsc{NoPerturbation}\}$
    \State Define $P_\text{video} \gets \{\textsc{AddNoise}, \textsc{Reverse}, \textsc{ReplaceWithIrrelevant}, \textsc{NoPerturbation}\}$
    \State Define $P_\text{sensor} \gets \{\textsc{AddJitter}, \textsc{ReplaceWithIrrelevant}, \textsc{NoPerturbation}\}$

    \If{RandomChoice([True, False])}
        \State $a' \gets$ RandomChoice($P_\text{audio}$)($a$)
    \Else
        \State $a' \gets a$
    \EndIf

    \If{RandomChoice([True, False])}
        \State $v' \gets$ RandomChoice($P_\text{video}$)($v$)
    \Else
        \State $v' \gets v$
    \EndIf

    \If{RandomChoice([True, False])}
        \State $s' \gets$ RandomChoice($P_\text{sensor}$)($s$)
    \Else
        \State $s' \gets s$
    \EndIf

    \State \Return $D' = \{a', v', s'\}$
\EndFunction
\end{algorithmic}
\end{algorithm*}

Cross-modal mismatch refers to the condition in which the semantic alignment between different input modalities—such as audio, video, and sensor streams—is disrupted. In real-world multi-modal systems, such mismatches frequently arise due to noise, missing data, or temporal desynchronization between modalities. Understanding and addressing cross-modal mismatch is crucial for building robust models capable of effective reasoning across modalities.

To systematically evaluate model robustness under such conditions, we introduce a synthetic cross-modal mismatch generation process. Given a clean multi-modal datapoint \( D = \{a, v, s\} \), where \( a \), \( v \), and \( s \) denote the synchronized audio, video, and sensor streams respectively, we construct a perturbed version \( D' = \{a', v', s'\} \) by applying one or more of the following perturbations:

\parlabel{Modality-Specific Noise Injection}: Gaussian or environmental noise is added to the audio \( a \) and/or video \( v \) streams, degrading signal fidelity while preserving temporal structure.

\parlabel{Temporal Reversal}: The temporal sequence of audio or video is reversed independently, altering the causal and sequential semantics of events.

\parlabel{Sensor Perturbation}: Random noise or jitter is added to sensor streams (e.g., IMU data), simulating faulty or low-resolution sensor readings.

\parlabel{Modal Replacement}: One or more modalities (e.g., audio) are replaced with semantically irrelevant counterparts sampled from other unrelated datapoints in the dataset, creating intentional cross-modal conflict.

These perturbations simulate realistic mismatches commonly encountered in egocentric and exocentric environments, such as microphone occlusion, corrupted video frames, or misaligned sensor logging. This synthetic mismatch generation enables controlled stress testing of multi-modal models, revealing their capacity to handle noisy, misaligned, or contradictory inputs across modalities. Algorithm \ref{alg_cross_modal_mismatch} explains the process used for generating cross-modal mismatch.


\section{Training and Implementation Details}
\label{supp:train_implement}

\subsection{Dataset for Multistage Training}
\label{supp_training_dataset}
\begin{table*}[!htb]
\centering
\caption{Datasets used at each training stage of \ourapproach. \dname contributes to all three stages, enabling both sensor-text alignment and robust fine-tuning under cross-modal mismatch.}
\vspace{-0.5em}
\label{tab_training_stage_dataset}
\resizebox{\linewidth}{!}{
\begin{tabular}{@{}lll|l@{}}
\toprule[2pt]
\multicolumn{2}{c}{\textbf{Training stage}} &
  \multicolumn{1}{c}{\textbf{Dataset}} &
  \multicolumn{1}{c}{\textbf{\#Pairs}} \\ \midrule[1.5pt]
\multicolumn{1}{l|}{\multirow{3}{*}{Modality-Text Pre-Training}} &
  \multicolumn{1}{l|}{Vision-Text} &
  \multicolumn{1}{l|}{\begin{tabular}[c]{@{}l@{}}InternVid-10M \cite{internvid10m}, WebVid-10M \cite{webvid10m}, \\ Panda-70M \cite{panda70m}, VIDAL-10M \cite{vidal10m}, \\ CC-3M \cite{cc3m}, DCI \cite{dci}\end{tabular}} &
  12.2 M \\
  \arrayrulecolor{shadecolor} \cmidrule[1pt](l){2-4}\arrayrulecolor{black}
\multicolumn{1}{l|}{} &
  \multicolumn{1}{l|}{Audio-Text} &
  \multicolumn{1}{l|}{WavCaps \cite{wavcaps}} &
  400K \\
    \arrayrulecolor{shadecolor} \cmidrule[1pt](l){2-4}\arrayrulecolor{black}
\multicolumn{1}{l|}{} &
  \multicolumn{1}{l|}{Sensor-Text} &
  \multicolumn{1}{l|}{OpenSQA \cite{llasa}, SensorCaps \cite{llasa}} &
  205K \\ \midrule[1.5pt]
\multicolumn{2}{l|}{Query-Token Alignment Joint-Training} &
  \multicolumn{1}{l|}{\begin{tabular}[c]{@{}l@{}}AVQA\cite{avqa}, AVSSD \cite{avssd}, \\ MUSIC-AVQA \cite{musicavqa}, \\ AVSD \cite{avsd}, \dname\end{tabular}} &
  403K \\ \midrule[1.5pt]
\multicolumn{2}{l|}{Modal-Discrepency Aware Fine-Tuning} &
  \begin{tabular}[c]{@{}l@{}}AVQA \cite{avqa}, AVSSD \cite{avssd}, \\ MUSIC-AVQA \cite{musicavqa}, \\ AVSD \cite{avsd}, \dname\end{tabular} &
  510K \\ \bottomrule[2pt]
\end{tabular}
}
\end{table*}
Along with our in-house data (\dname), we use publicly available datasets to train the video, audio, and sensor encoders. To pre-train the sensor encoder, we use epic kitchen \cite{epic-kitchen}, ego4D \cite{ego4d},HHAR~\cite{stisen2015smart}, UCI-HAR~\cite{reyes2016transition}, Shoaib~\cite{shoaib2014fusion}, MotionSense~\cite{malekzadeh2019mobile}, PAMAP2 \cite{roggen2010collecting} data. We use pre-trained SigLIP as our video encoder and then fine-tune it with datasets from videoLLama2 \cite{videollama2}. Similarly, we use a pre-trained audio encoder, Beats, and fine-tune it with WavCaps \cite{wavcaps} datasets \cite{beats}. We leverage SensoCaps and OpenSQA~\cite{llasa} for the sensor pretraining part. Table \ref{tab_training_stage_dataset} summarizes the dataset used at different stages of training.

\label{supp_implementation}
\subsection{Hyperparameters for Training}
\label{supp_hyperparams}

\ourapproach has 8.5B parameters, including all the encoders, projection layers, \amodule, and LLM backbone. Table~\ref{tab_hyperparams} summarizes the key hyperparameters used during training. 

\begin{table}[!htb]
\caption{Key hyperparameters used in training \ourapproach. Token counts reflect the number of input tokens per modality. We adopt a 6-layer transformer with 8 attention heads, a LoRA rank of 4256, and use AdamW for optimization.}
\label{tab_hyperparams}
\resizebox{\linewidth}{!}{
\begin{tabular}{l|c|r}
\toprule[2pt]
Description                                  & Notation & \multicolumn{1}{l}{Value} \\ \midrule[1pt]
\multicolumn{1}{l|}{Number of audio tokens}  & $\mathbf{L}_a$      & 1496                      \\
\multicolumn{1}{l|}{Number of video tokens}  & $\mathbf{L}_v$      & 1352                      \\
\multicolumn{1}{l|}{Number of sensor tokens} & $\mathbf{L}_s$      & 120                       \\
\multicolumn{1}{l|}{Embedding dimension}    & $\mathbf{E}$        & 3584                      \\
\multicolumn{1}{l|}{Number of total token}   & $\mathbf{L}$        & 2968                      \\
\multicolumn{1}{l|}{Numer of heads}          & $\mathbf{h}$        & 8                         \\
\multicolumn{1}{l|}{Number of encoder layer} & $\mathbf{N}$        & 6                         \\ 
Each head dimension                         & $\mathbf{d}_k$      & 448                       \\ 
Batch size (local/global)                         & -      &  1/4                       \\ 
LoRA rank                         & $\mathbf{r}$      & 4256                     \\ 
Optimizer                        & -    & AdamW                      \\ 
Weight decay                        &  -      & 0.03                       \\ 
\bottomrule[2pt]
\end{tabular}
}
\end{table}

\subsection{Train-Test split}
\label{supp_train_test_split}

For all publicly available datasets used during pre-training and fine-tuning, we adopt the official train--test splits provided by their respective authors. For our curated dataset, \dname, we create a standardized train--test split to ensure consistent evaluation and reproducibility. To prevent data leakage and overfitting, we ensure the input sessions for curating \dname train and test split remain completely separated. The split files are publicly available in our GitHub repository \textcolor{cvprblue}{\url{https://github.com/BASHLab/RAVEN/tree/main/avs-qa-dataset}}.



\section{Evaluation Details}
\subsection{Evaluation Baselines}
\label{supp_baselines}

\parlabel{Video-LLaMA} 
Video-LLaMA extends LLaMA by incorporating frozen video encoders (TimeSformer, X-CLIP) to extract spatio-temporal features, which are linearly projected into the LLM input space. It is trained via instruction tuning and multi-modal supervised learning, enabling video captioning, question answering, and reasoning with generalization from few-shot examples.

\parlabel{Video-LLaMA2} Video-LLaMA-2 builds upon its predecessor by introducing spatio-temporal connectors, which better align video representations with the LLM input through a more structured fusion mechanism. Additionally, Video-LLaMA-2 leverages more powerful video encoders and larger training corpora, making it more robust for real-world multimodal applications.

\parlabel{PandaGPT} PandaGPT integrates CLIP for visual features and BEATs for audio features, followed by a Q-Former to project them into the token space of a language model (Vicuna). PandaGPT supports multi-turn dialogue grounded in both visual and auditory content, enabling it to reason over video-audio-text contexts.

\parlabel{Macaw-LLM} Macaw-LLM adopts a modular design where a dedicated encoder process each modality, and the features are fused into a shared embedding space for the language model. Inspired by BERT-style pretraining, Macaw-LLM supports tasks such as cross-modal retrieval, multimodal classification, and audio-visual QA. 

\parlabel{VideoChat} VideoChat introduces a video-grounded dialogue system that enables interactive conversations about dynamic visual content. It uses a pre-trained video encoder (like X-CLIP or SwinBERT) to extract frame-wise representations and then aligns these with LLaMA through lightweight adapters. VideoChat supports both single-turn and multi-turn video QA, offering real-time conversational abilities over video inputs. It was among the first open-source models to demonstrate effective temporal video grounding in LLM-based dialogue.

\parlabel{VideoChatGPT} VideoChatGPT extends VideoChat by incorporating end-to-end video-LM alignment with improved temporal reasoning and multi-frame understanding. It utilizes a stronger video encoder and enhanced fusion modules (e.g., spatio-temporal attention layers) to feed richer video context into the LLM. 

\parlabel{VALLEY} VALLEY (VisuAL Langauge Learner with Large memorY) is designed for multi-modal memory-augmented video reasoning. It focuses on long-term memory alignment across video segments and text, allowing the model to retain and reference past frames effectively during reasoning. VALLEY combines a hierarchical visual encoder with a memory-enhanced transformer decoder that interacts with a language model, enabling it to handle long videos and multi-step reasoning tasks such as procedural understanding, storytelling, and temporal localization.

\parlabel{VTimeLLM} VTimeLLM (Video-Time Language Model) focuses on temporal video understanding by aligning spatio-temporal features with natural language in a query-aware manner. It introduces a temporal reasoning module that captures the order, duration, and causality of events in video segments. Using a dual-stream architecture with temporal attention and frame-level token sampling, VTimeLLM fuses visual and language information for downstream tasks such as video QA, moment retrieval, and video narration.

\parlabel{AV-LLM} AV-LLM integrates auditory and visual modalities using CLIP for images/videos and Whisper or BEATs for audio with a frozen LLaMA. It employs a cross-modal projection layer and lightweight adapters to fuse the modalities, enabling zero-shot and instruction-tuned tasks like audio-visual QA, event description, and sound-source reasoning. 

\parlabel{AVicuna} AViCuna is a chat-centric audio-visual instruction-following model that combines audio and video features into a unified token stream for a conversational LLM based on Vicuna. It uses Q-Former modules to encode BEATs for audio and CLIP for video features, and feeds these to the LLM via a learned query-token bridge. 

\parlabel{OpenFlamingo} OpenFlamingo fuses a frozen CLIP-ViT with a pre-trained language model via a perceiver-style cross-attention module. The key innovation lies in its interleaved visual-text token interface, which allows the model to reason over multimodal sequences without further fine-tuning. OpenFlamingo supports tasks such as image captioning, VQA, and multi-image reasoning in an efficient and instruction-following setting.

\parlabel{SahreGPT4V} ShareGPT4V emphasizes the importance of caption quality in multimodal learning, showing that even a modest amount of rich, semantically dense image-text pairs can significantly improve LMM performance. It uses GPT-4V to generate 100k captions and further extend the dataset to a 1.2m sample by using a caption model. ShareGPT4V is then fine-tuned with this caption dataset as a foundational MMLLM.

\parlabel{MiniGPT-4} MiniGPT-4 mimics GPT-4V's capabilities using open components. It pairs a frozen CLIP-ViT with a Vicuna-based LLM via a linear projection layer, trained with a two-stage instruction tuning pipeline. MiniGPT-4 achieves strong performance with low computational cost.

\parlabel{BLIP-2.6} BLIP-2.6 is an evolution of BLIP-2, further improving the alignment between vision encoders and LLMs using a multistage pretraining and fine-tuning strategy. It enhances the Q-Former mechanism and supports longer and denser vision-language interactions with better grounding fidelity. BLIP-2.6 shows improvements in instruction following, fine-grained captioning, and long-context multimodal tasks while maintaining the zero-shot generalization strength of BLIP-2.

\parlabel{InstructBLIP}
InstructBLIP is an instruction-tuned extension of the BLIP-2 family, designed to align vision-language pretraining with task-specific prompts. It introduces a flexible prompting mechanism and uses a frozen vision encoder with a trainable Q-Former to bridge the modality gap to an LLM.

\subsection{Evaluation Datasets}
\label{supp:val_dataset}
\parlabel{InternVid-10M}
InternVid-10M is a large-scale video-text dataset comprising approximately 10 million video-caption pairs, designed to support pretraining of multimodal large language models. The videos are sourced from diverse domains, and the captions are refined to improve visual-textual alignment.

\parlabel{WebVid-10M}
WebVid-10M consists of 10 million video-text pairs harvested from web sources, particularly short-form videos with associated metadata or alt-text. Although noisier than manually curated datasets, its sheer scale makes it valuable for video-language pretraining.

\parlabel{Panda-70M}
Panda-70M is a massive multimodal dataset containing over 70 million aligned video, audio, and text triplets. It is curated from open-domain videos, including instructional content, to cover a wide variety of real-world scenarios. The dataset is designed for training models that require joint understanding of video, audio, and language, enabling tasks such as multimodal reasoning, audio-visual captioning, and cross-modal retrieval at scale.

\parlabel{Vidal-10M}
VIDAL-10M is a curated dataset comprising 10 million high-quality video-caption pairs aimed at enhancing temporal and contextual understanding in multimodal models. It includes dense and descriptive captions aligned with diverse video domains, enabling robust pretraining for video-language models. VIDAL-10M emphasizes temporal consistency and semantic diversity, supporting tasks like video QA, moment retrieval, and event understanding.

\parlabel{CC-3M}
CC-3M is a widely-used image-text dataset containing approximately 3 million image-caption pairs sourced from the web. The captions are filtered and cleaned alt-text annotations that loosely describe the visual content. While the descriptions can be noisy and lack fine-grained detail, it is valuable for large-scale vision-language pretraining, especially for image-text retrieval, captioning, and contrastive representation learning.

\parlabel{DCI}
DCI is a dataset developed to improve instruction-following in vision-language models by pairing images with rich, instruction-style descriptions. The captions are generated using large language models guided by carefully designed prompts to increase informativeness and task relevance. DCI serves as a bridge between standard image-caption datasets and instruction-tuned models, supporting applications like visual instruction-following, grounded question answering, and image-based reasoning.

\parlabel{WavCaps}
WavCaps is a large-scale audio-text dataset designed to enhance audio-language pretraining. It includes over 400,000 audio clips paired with captions, either collected from metadata or generated via model-based annotation pipelines. Covering a wide range of sound events—from speech and music to environmental and mechanical sounds—WavCaps supports tasks such as audio captioning, sound event detection, and cross-modal audio-text retrieval.

\parlabel{SensorCaps}
SensorCaps is a pioneering sensor-language dataset that pairs time-series data from inertial measurement units (IMUs) and other body-worn sensors with detailed natural language descriptions. Designed to support tasks like sensor captioning and multimodal grounding, SensorCaps bridges wearable sensing data with large language models. It enables multimodal LLMs to reason about human actions, physical context, and temporal dynamics from sensor inputs.

\parlabel{OpenSQA}
OpenSQA is a benchmark dataset for sensor-based question answering, aiming to bring structured reasoning capabilities to models processing sensor time-series data. It includes labeled QA pairs grounded in sensor streams from IMU collected in real-world contexts. OpenSQA supports open-ended and multiple-choice questions, making it a valuable testbed for evaluating sensor-to-text alignment and semantic understanding in multimodal models.

\parlabel{AVSD}
AVQA is a benchmark dataset specifically designed for evaluating audio-visual reasoning capabilities in multimodal models. It includes videos paired with open-ended and multiple-choice questions that require joint analysis of both visual content and audio cues. AVQA challenges models to perform fine-grained audio-visual fusion for answering questions about actions, events, or contextual elements that span both modalities.

\parlabel{AVSSD}
AVSSD is a large-scale dataset containing over 200,000 audio-video clips spanning 310 sound classes. Each clip is approximately 10 seconds long and is sourced from YouTube, covering a wide range of natural and human-made sounds. AVSSD supports weakly-supervised learning and cross-modal modeling, especially for tasks like sound classification, audio-visual event detection, and audio grounding in video.

\parlabel{MUSIC-AVQA}
MUSIC-AVQA is a specialized dataset designed for audio-visual question answering in musical contexts, where questions require understanding of both the visual performance and the auditory output of musical instruments. It is built upon the MUSIC dataset, which includes isolated instrument performances. MUSIC-AVQA extends MUSIC with over 7,000 QA pairs involving tasks such as instrument identification, sound localization, source counting, and event timing. The questions are crafted to assess fine-grained audio-visual reasoning, where answers depend on spatial, temporal, and semantic alignment of what is seen and heard. 

\parlabel{AVQA} AVQA is a benchmark dataset specifically designed for evaluating audio-visual reasoning capabilities in multimodal models. It includes videos paired with open-ended and multiple-choice questions that require joint analysis of both visual content and audio cues. AVQA challenges models to perform fine-grained audio-visual fusion for answering questions about actions, events, or contextual elements that span both modalities.

\parlabel{EgoThink}
EgoThink is a benchmark designed to evaluate the first-person perspective reasoning capabilities of vision-language models (VLMs). It comprises question-answer pairs derived from egocentric video clips, focusing on six core capabilities across twelve detailed dimensions. The dataset emphasizes tasks that require models to understand and reason from a first-person viewpoint, such as anticipating future actions or interpreting personal experiences. Evaluations of eighteen popular VLMs on EgoThink reveal that, while models like GPT-4V perform well in certain areas, there remains significant room for improvement in first-person perspective tasks. EgoThink serves as a valuable resource for advancing research in embodied artificial intelligence and robotics.

\subsection{Evaluation Metric}
\label{supp:eval_metric}
\begin{figure}
    \centering
    \includegraphics[width=\linewidth]{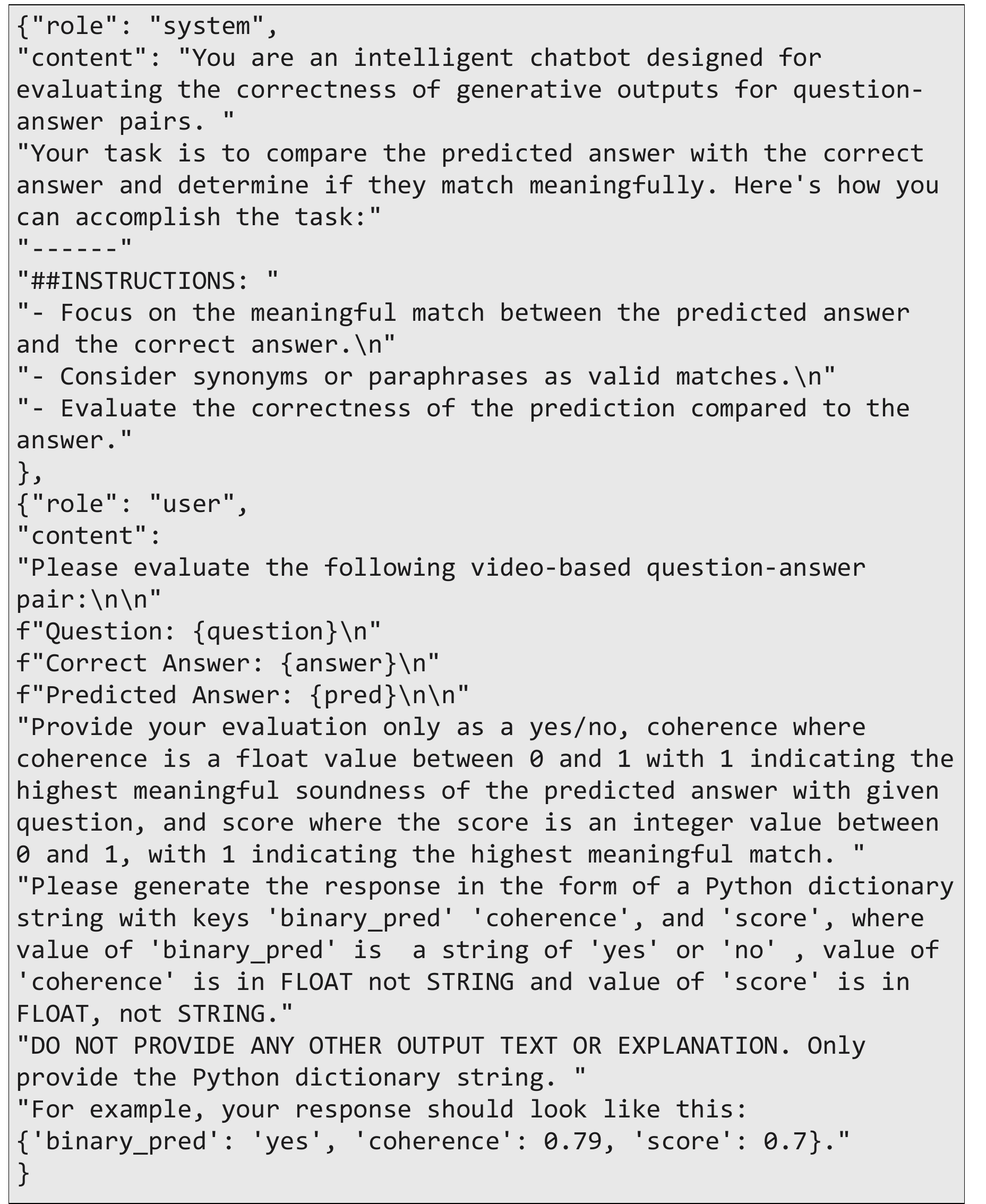}
    \caption{System and user prompt used to evaluate the generated answer quality.}
    \label{eval_prompt}
\end{figure}

Following previous work ~\cite{videochatgpt}, we leverage GPT-3.5-turbo to evaluate the generated answer quality. Figure \ref{eval_prompt} depicts the evaluation prompt. 
\section{Ablation Study}
\label{supp_ablation_study}

\begin{table}[!htb]
\caption{Comparison of video encoders across three QA benchmarks. SigLIP consistently outperforms all ViT variants, demonstrating stronger temporal and visual grounding for video-based question answering.}
\label{tab_ve_ablation}
\resizebox{\linewidth}{!}{
\begin{tabular}{l|ccc}
\toprule[2pt]
\multicolumn{1}{c|}{} & \multicolumn{3}{c}{\textbf{Datasets}} \\ \cmidrule(l){2-4}
\multicolumn{1}{c|}{\multirow{-2}{*}{\textbf{\begin{tabular}[c]{@{}c@{}}Video\\ Encoder\end{tabular}}}} &
  \multicolumn{1}{c}{\textbf{\begin{tabular}[c]{@{}c@{}}MSVD- \\ QA\end{tabular}}} &
  \multicolumn{1}{c}{\textbf{\begin{tabular}[c]{@{}c@{}}MSRVTT- \\ QA\end{tabular}}} &
  \multicolumn{1}{c}{\textbf{\begin{tabular}[c]{@{}c@{}}ActivityNet- \\ QA\end{tabular}}} \\ \midrule[1.5pt]
ViT-B/16              & 65.7        & 51.4       & 45.9       \\
ViT-L/14              & 67.3        & 53.7       & 47.2       \\
ViT-H/14              & 67.5        & 54.2       & 47.5       \\ \midrule
\rowcolor[HTML]{DAE8FC} 
SigLip                & 73.3        & 63.1       & 57.6       \\ \bottomrule[2pt]
\end{tabular}
}
\end{table}
\begin{table}[!htb]
\caption{Performance of audio encoders across QA datasets. BEATs achieves the highest accuracy on all benchmarks, surpassing Whisper variants in multimodal reasoning tasks.}
\label{tab_ae_ablation}
\resizebox{\linewidth}{!}{
\begin{tabular}{l|ccc}
\toprule[2pt]
\multicolumn{1}{c|}{} & \multicolumn{3}{c}{\textbf{Datasets}} \\ \cmidrule(l){2-4} 
\multicolumn{1}{c|}{\multirow{-2}{*}{\textbf{\begin{tabular}[c]{@{}c@{}}Audio\\ Encoder\end{tabular}}}} &
  \multicolumn{1}{c}{\textbf{\begin{tabular}[c]{@{}c@{}}MSVD- \\ QA\end{tabular}}} &
  \multicolumn{1}{c}{\textbf{\begin{tabular}[c]{@{}c@{}}MSRVTT- \\ QA\end{tabular}}} &
  \multicolumn{1}{c}{\textbf{\begin{tabular}[c]{@{}c@{}}ActivityNet- \\ QA\end{tabular}}} \\ \midrule[1.5pt]
Whisper-T          & 66.5       & 51.6       & 46.2 \\
Whisper-B          & 67.7       & 53.1       & 47.4 \\
Whisper-S         & 68.1       & 53.9       & 47.6 \\ \midrule
\rowcolor[HTML]{DAE8FC} 
BEATs                 & 73.3       & 63.1       & 57.6 \\ \bottomrule[2pt]
\end{tabular}
}
\end{table}

\begin{figure}[!htb]
    \centering
    \includegraphics[width=0.9\linewidth]{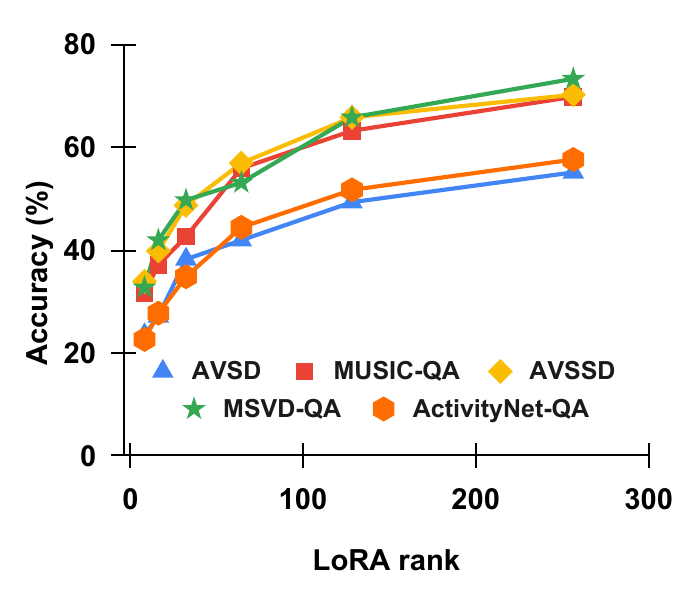}
    \vspace{-1em}
    \caption{Impact of LoRA rank on QA accuracy across five benchmarks. Accuracy improves steadily with higher ranks, saturating near 256, indicating that moderate-rank adapters suffice for effective multimodal alignment and reasoning.}
    \vspace{-1em}
    \label{lora_rank}
\end{figure}

\begin{table*}[!htb]
\centering
\caption{Effect of Frame selection strategy on \amodule.}
\label{supp_tab_frame_selection}
\resizebox{0.7\linewidth}{!}{
\begin{tabular}{lcccccc}
\toprule[2pt]
\multicolumn{1}{l|}{\textbf{\begin{tabular}[c]{@{}c@{}}Sampling \\ Method\end{tabular}}}  &
  \multicolumn{1}{c}{\textbf{AVSD}} &
  \multicolumn{1}{c}{\textbf{\begin{tabular}[c]{@{}c@{}}MUSIC- \\ QA\end{tabular}}} &
  \multicolumn{1}{c}{\textbf{AVSSD}} &
  \multicolumn{1}{c}{\textbf{\begin{tabular}[c]{@{}c@{}}MSVD- \\ QA\end{tabular}}} &
  \multicolumn{1}{c}{\textbf{\begin{tabular}[c]{@{}c@{}}MSRVTT- \\ QA\end{tabular}}} &
  \multicolumn{1}{c}{\textbf{\begin{tabular}[c]{@{}c@{}}ActivityNet- \\ QA\end{tabular}}} \\ \midrule[1.5pt]
\multicolumn{1}{l|}{Random}       & 45.2 & 55.6 & 49.3 & 50.7 & 43.1 & 47.2 \\
\multicolumn{1}{l|}{Fixed Stride} & 54.9 & 69.3 & 70.1 & 72.9 & 62.8 & 57.4 \\
\multicolumn{1}{l|}{Uniform}     & 55.1 & 69.8 & 70.2 & 73.3 & 63.1 & 57.6 \\
\multicolumn{1}{l|}{Oracle}                           & 55.2 & 70.1 & 70.2 & 73.4 & 63.5 & 57.6 \\ \bottomrule[2pt]
\end{tabular}
}
\end{table*}
\parlabel{Effect of Modality Encoder} We investigate the influence of visual and audio encoder choices on model performance across three video QA benchmarks (Tables \ref{tab_ve_ablation}, \ref{tab_ae_ablation}). For vision, scaling standard ViT architectures from B/16 to H/14 yields only marginal improvements (e.g., +1.8\% on MSVD-QA), suggesting limited benefits from increasing model capacity alone. In contrast, substituting ViT with SigLip, a vision-language pretrained model leads to substantial performance gains (73.3 vs. 67.5 on MSVD-QA), demonstrating the importance of cross-modal alignment during pretraining. 
On the audio side, scaling Whisper encoders from Tiny to Small results in modest improvements (e.g., +1.6\% on MSVD-QA), but all Whisper variants are outperformed by BEATs, a model pretrained on diverse acoustic signals. Notably, BEATs achieves a +5.2\% gain over Whisper-Small on MSVD-QA, highlighting the efficacy of domain-specific audio pertaining.

\parlabel{LoRA Rank Selection} Figure~\ref{lora_rank} shows an ablation on LoRA rank. Lower ranks improve efficiency but may limit representational capacity, while higher ranks offer greater adaptability at a higher cost. Performance peaks at $r = 256$, indicating it provides the best trade-off between computational overhead and task effectiveness.

\begin{table}[!htb]
\caption{Comparison of \amodule with General Fusion Approaches. \amodule performs better due to its token-level reasoning capabilities.}
\vspace{-0.75em}
\label{tab_genfuse_ablation}
\centering
\resizebox{0.7\linewidth}{!}{
\begin{tabular}{l|cc}
\toprule[2pt]
\multicolumn{1}{c|}{} & \multicolumn{2}{c}{\textbf{Datasets}} \\ \cmidrule(l){2-3}
\multicolumn{1}{c|}{\multirow{-2}{*}{\textbf{\begin{tabular}[c]{@{}c@{}}Fusion \\ Model\end{tabular}}}} &
  \multicolumn{1}{c}{\textbf{\begin{tabular}[c]{@{}c@{}}AVSSD\end{tabular}}} &
  \multicolumn{1}{c}{\textbf{\begin{tabular}[c]{@{}c@{}}MSRVTT- \\ QA\end{tabular}}}  \\ \midrule[1.5pt]
Imagebind             & 27.8        & 27.8          \\
MBT              & 64.1        & --         \\
AVFIC              & --        & 19.4           \\ \midrule
\rowcolor[HTML]{DAE8FC} 
\amodule                & \textbf{70.2}        & \textbf{63.1}           \\ \bottomrule[2pt]
\end{tabular}
\vspace{-1em}
}
\end{table}
\parlabel{Comparison of \amodule with General Fusion Approaches}
We compare \amodule with state-of-the-art general-purpose fusion models (ImageBind~\cite{imagebind}, MBT~\cite{nagrani2021attention}, and AVFIC~\cite{nagrani2022learning}), which are not optimized for QA tasks. As shown in Table~\ref{tab_genfuse_ablation}, \amodule outperforms these models, highlighting the benefit of QA-specific supervision and token-level fusion for effective reasoning.

\parlabel{Effect of Frame Selection Strategy} We adopted a uniform frame sampling strategy, consistent with prior video QA and egocentric video research \cite{videollama2, avicuna}, to ensure fair comparison and reproducibility.

However, we evaluated \ourapproach leveraging several alternative frame selection methods, including fixed stride, random, and oracle-based selection. Notably, performance (Table \ref{supp_tab_frame_selection}) remained mostly consistent across all methods, except for random sampling, which caused increased variability and occasional performance drops. For oracle-based selection, we used salient frame annotations and metadata from EpicKitchen-100 and Ego4D to choose visually informative frames. While this led to small improvements in some vision-intensive queries, the overall trends of \ourapproach’s performance remained unchanged. These findings suggest that \amodule is resilient to the choice of frame selection, although future work could investigate learned or adaptive methods to further enhance performance on long video reasoning.

\section{Compute Cost and Environmental Impact}
\label{supp_cost}

We train our model using four NVIDIA A100 GPUs (80GB each) with a total CPU memory of 256GB. Evaluation is performed on four NVIDIA L40S GPUs (46GB each). Training runs for 120 hours with a local batch size of 1 and a global batch size of 4. We use a learning rate of $1\times10^{-3}$ for the projection layers and $1\times10^{-5}$ for fine-tuning the encoder layers.

We estimate the total energy consumption to be approximately 1,200 kWh, based on the average power draw of an A100 system under mixed precision load. Following the ML CO$_2$ emissions calculator~\cite{lacoste2019quantifying}, this corresponds to an estimated carbon footprint of 420 kgCO$_2$e when using the U.S. average energy mix.


\begin{figure*}[!htb]
    \centering  
    \includegraphics[width=\linewidth]{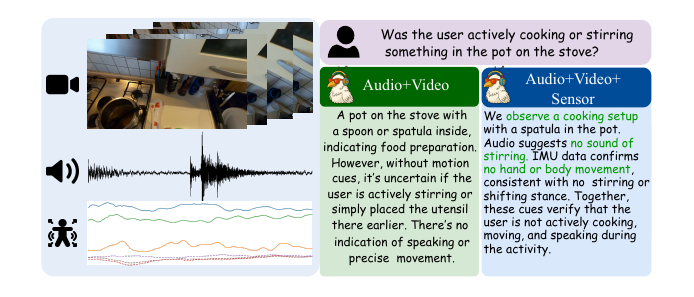}
    \caption{Example illustrating the value of sensor input for activity disambiguation. Given the question \textit{“Was the user actively cooking or stirring something in the pot on the stove?”}, the \textbf{Audio+Video} model observes a cooking scene but cannot confirm active engagement due to the absence of motion cues. In contrast, the \textbf{Audio+Video+Sensor} model leverages IMU data to detect a lack of body movement and integrates audio signals to confirm no stirring, allowing it to infer that the user is \textbf{not actively cooking}.
}
    \label{fig_qual_av}
\end{figure*}

\begin{figure*}[!htb]
    \centering
    \includegraphics[width=\linewidth]{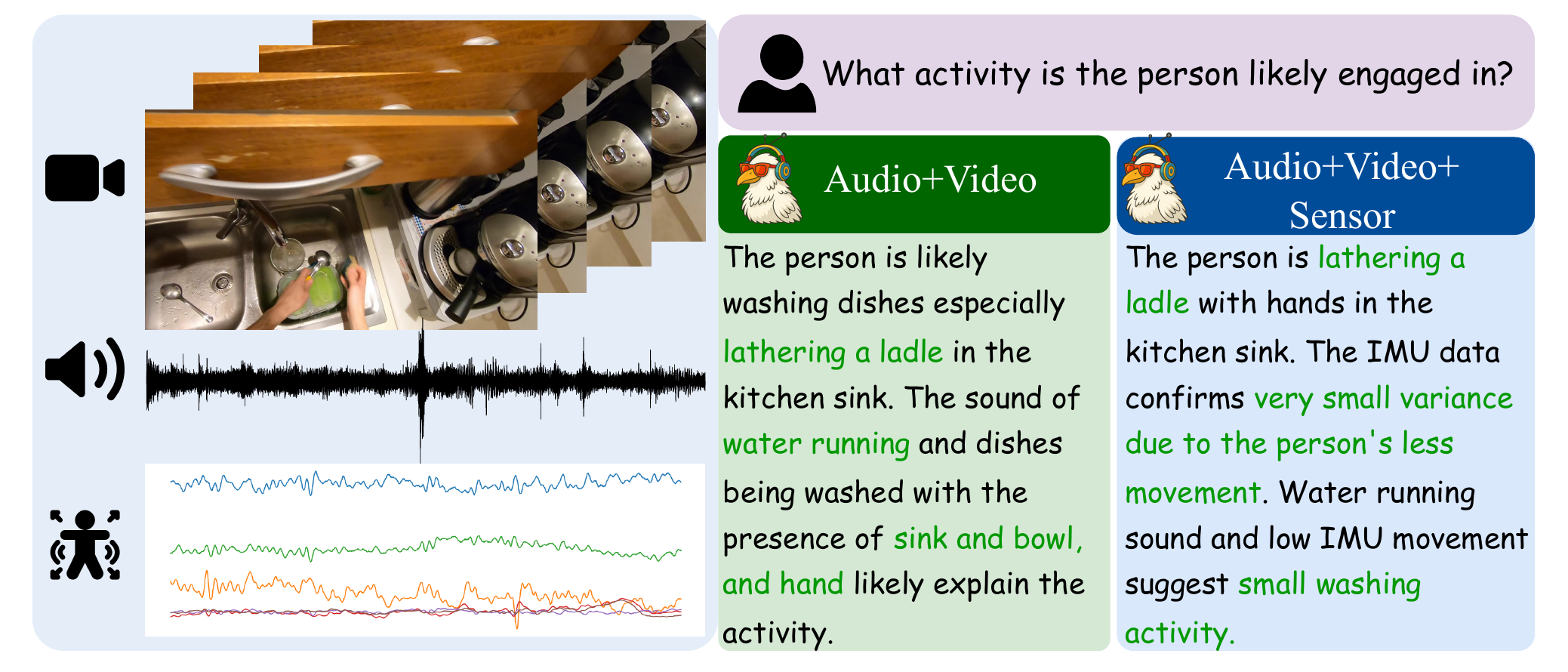}
    \caption{Example illustrating subtle activity disambiguation using multimodal reasoning. Given the question \textit{“What activity is the person likely engaged in?”}, the \textbf{Audio+Video} model identifies dishwashing activity based on sink visibility and audio cues such as water flow. The \textbf{Audio+Video+Sensor} model enhances this understanding by incorporating IMU data, which reveals \textbf{low hand and body movement}. This confirms a controlled, repetitive action consistent with small-scale washing (e.g., lathering a ladle), demonstrating the added value of sensor input for refining temporal and motion-level interpretations.
}
    \label{fig:enter-label}
\end{figure*}

\section{Qualitative Results}
\label{supp_qualitative}
\begin{figure*}[!htb]
    \centering
    \includegraphics[width=\linewidth]{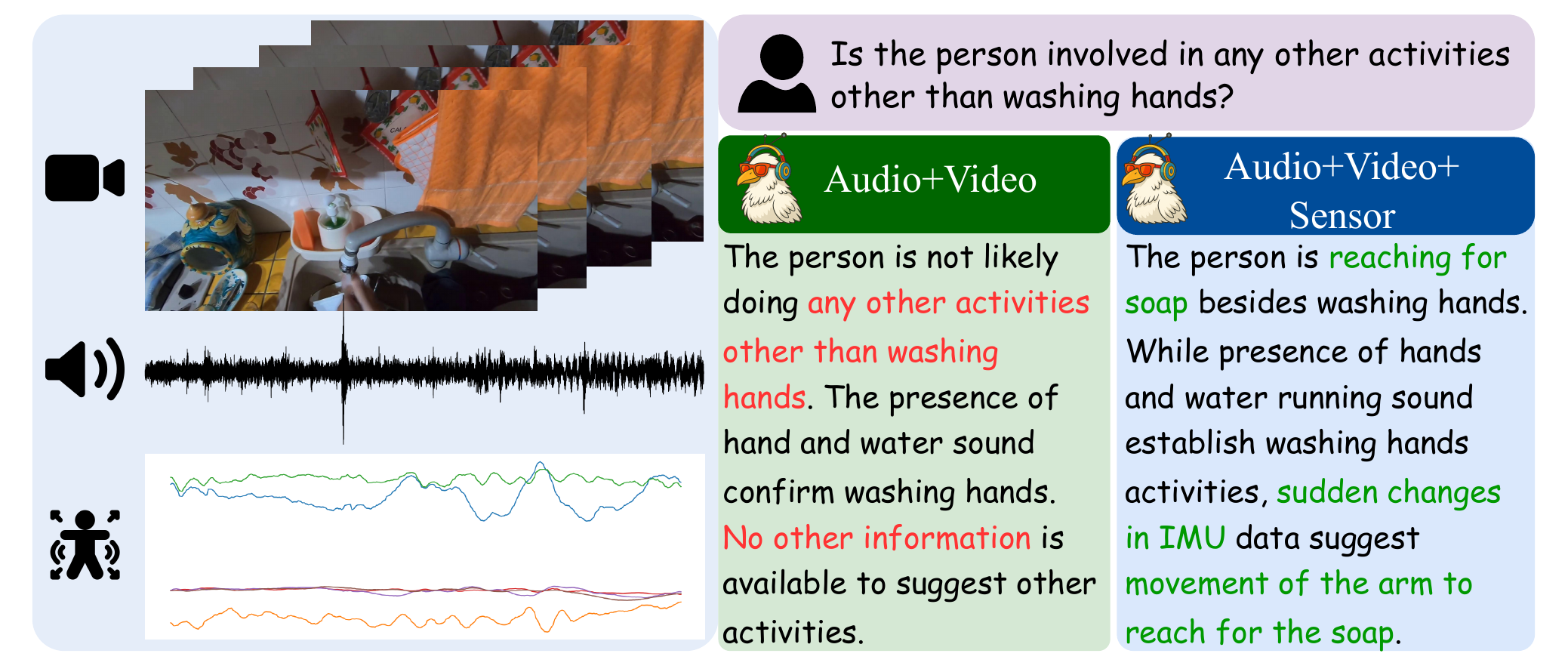}
    \caption{
Example demonstrating the added value of sensor data in identifying subtle concurrent actions. Given the question \textit{“Is the person engaged in any other activities other than washing hands?”}, the \textbf{Audio+Video} model detects only hand presence and water sounds, concluding that no other activities are evident. In contrast, the \textbf{Audio+Video+Sensor} model \textbf{identifies a sudden IMU spike, indicating arm movement associated with reaching for soap--capturing} a secondary action that is visually and acoustically ambiguous.
}
    \label{fig:qual_soap}
\end{figure*}

\begin{figure*}[!htb]
    \centering
    \includegraphics[width=\linewidth]{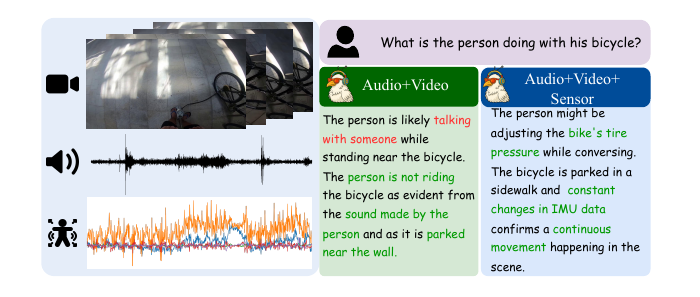}
    \caption{
Example showcasing multimodal reasoning for fine-grained activity understanding. Given the question \textit{“What is the person doing with his bicycle?”}, the \textbf{Audio+Video} model identifies that the person is not riding the bicycle and is likely talking nearby. In contrast, the \textbf{Audio+Video+Sensor} model captures \textbf{continuous IMU fluctuations, suggesting active engagement}, such as adjusting the bike’s tire pressure, demonstrating the added interpretive power of sensor input.
}
    \label{fig:qual_tire}
\end{figure*}

\begin{figure*}[htb]
    \centering
    \vspace{-33em}\includegraphics[width=\linewidth]{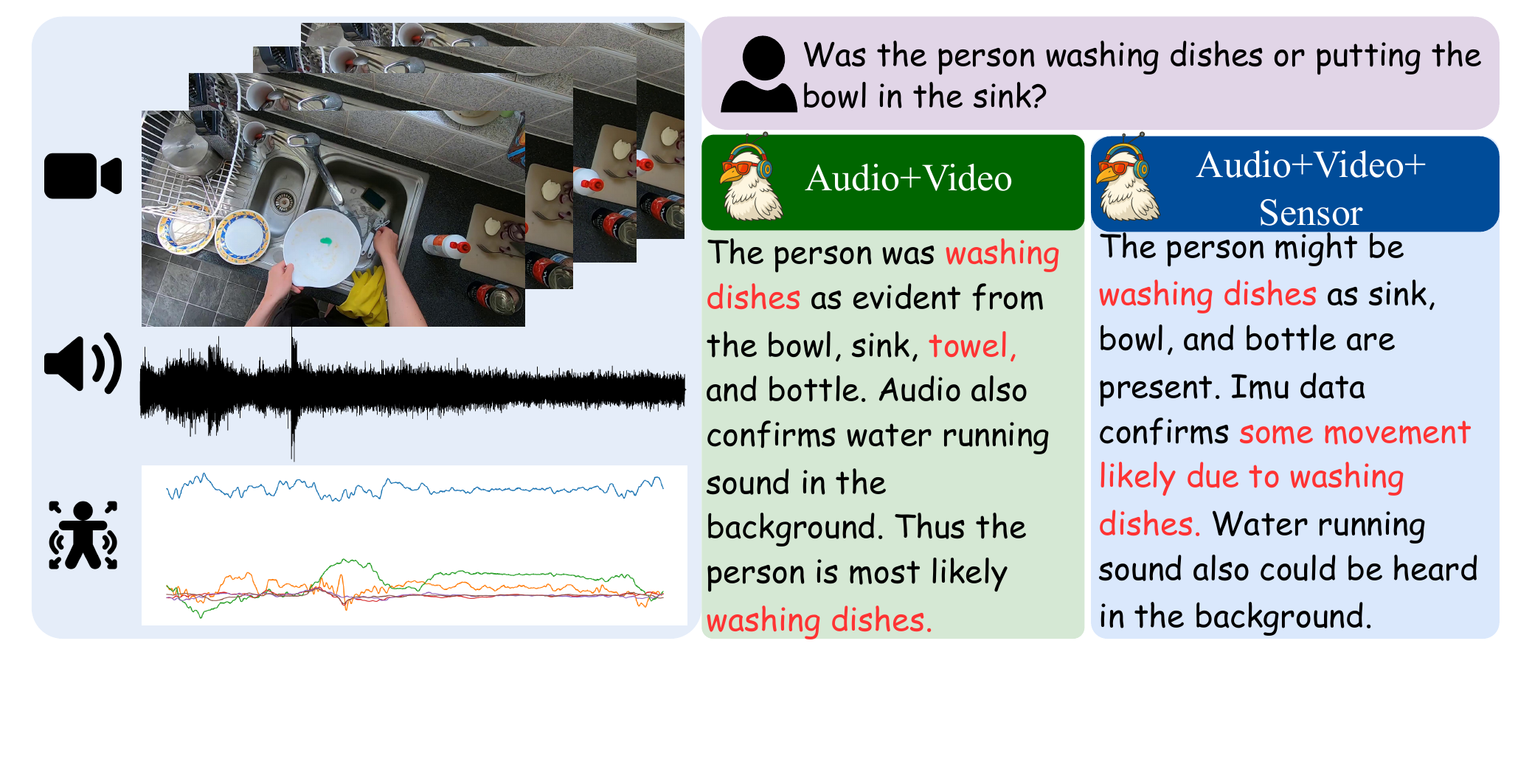}
    \caption{
Example illustrating confirmatory reasoning across modalities. Given the question \textit{“Was the person washing dishes or putting the bowl in the sink?”}, the \textbf{Audio+Video} model infers dishwashing based on visible objects (bowl, sink, towel) and background water sounds. The \textbf{Audio+Video+Sensor} model tries to strengthen this conclusion with \textbf{IMU evidence from the wrong source, inconsistent with washing actions}, reinforcing the activity label through motion-based verification.
}
    \label{fig:qual_bad}
\end{figure*}

Figures~\ref{fig_qual_av} -- \ref{fig:qual_bad} illustrate the performance of \ourapproach across diverse real-world scenarios. While \ourapproach demonstrates strong performance using only audio and visual inputs, the inclusion of sensor data consistently improves robustness and interpretability.

In particular, \ref{fig:qual_soap} and \ref{fig:qual_tire} highlight how sensor information enhances the correctness and relevance of both the predicted answer and its supporting explanation. Conversely, Figure~\ref{fig:qual_bad} presents a failure case where the model, even with full audio-video-sensor input, fails to infer the correct task due to subtle contextual clues across modalities that might not clearly differentiate similar tasks, hindering accurate inference. 
\balance


\clearpage

\end{document}